\documentclass[10pt,journal]{IEEEtran}

\expandafter\let\csname equation*\endcsname=\relax

\expandafter\let\csname endequation*\endcsname=\relax

\usepackage{amsmath,amssymb,amsthm}

\usepackage{tgpagella}
\usepackage{amsopn}
\usepackage{times}
\usepackage{graphicx}
\usepackage{nopageno}
\usepackage{eso-pic}

\usepackage[ruled,vlined]{algorithm2e}

\usepackage{relsize}
\usepackage{cases}

\usepackage{diagbox}
\usepackage{tabularx}
\usepackage{pifont}
\usepackage{enumerate} 
\usepackage{color}
\usepackage{fancyhdr}
\usepackage[justification=centering,textfont={small}]{caption}
\SetKwRepeat{Do}{do}{until}

\usepackage{cite}
\usepackage{textcomp}
\usepackage{array}
\usepackage{xcolor}
\usepackage{bm}
\usepackage{booktabs}

\usepackage{multirow}
\usepackage{epstopdf}
\usepackage{amsthm}

\usepackage[subfigure]{graphfig}
\def\BibTeX{{\rm B\kern-.05em{\sc i\kern-.025em b}\kern-.08em
    T\kern-.1667em\lower.7ex\hbox{E}\kern-.125emX}}

\begin{document}

\title{CityGPT: Towards Urban IoT Learning, Analysis and Interaction with Multi-Agent System}
\author{
Qinghua~Guan,~Jinhui~Ouyang,~Di~Wu~\IEEEmembership{Member,~IEEE},~Weiren~Yu,~\IEEEmembership{Member,~IEEE}
\thanks{J. Ouyang, Yi. Zhu, Xiang Yuan, and D. Wu are with the Key Laboratory for Embedded and Network Computing of Hunan Province, Hunan University, Changsha, Hunan 410082, China (e-mail: \{oldyoung, zyj1936013472, yuanxiang, dwu\}@hnu.edu.cn). W. Yu is with the Department of Computer Science, University of Warwick, Coventry CV4 7AL, UK (e-mail: weiren.yu@warwick.ac.uk).} 
}

\maketitle

\begin{abstract}
The spatiotemporal data generated by massive sensors in the Internet of Things (IoT) is extremely dynamic, heterogeneous, large scale and time-dependent. It poses great challenges ({\em e.g.} accuracy, reliability, and stability) in real-time analysis and decision making for different IoT applications. The complexity of IoT data prevents the common people from gaining a deeper understanding of it. Agentized systems help address the lack of data insight for the common people. We propose a generic framework, namely CityGPT, to facilitate the learning and analysis of IoT time series with an end-to-end paradigm. CityGPT employs three agents to accomplish the spatiotemporal analysis of IoT data. The requirement agent facilitates user inputs based on natural language. Then, the analysis tasks are decomposed into temporal and spatial analysis processes, completed by corresponding data analysis agents (temporal and spatial agents). Finally, the spatiotemporal fusion agent visualizes the system's analysis results by receiving analysis results from data analysis agents and invoking sub-visualization agents, and can provide corresponding textual descriptions based on user demands. To increase the insight for common people using our framework, we have agnentized the framework, facilitated by a large language model (LLM), to increase the data comprehensibility. Our evaluation results on real-world data with different time dependencies show that the CityGPT framework can guarantee robust performance in IoT computing.

\end{abstract}


       



\begin{IEEEkeywords}
Transportation Forecasting; AI Agent System; Spatial-temporal Data.
\end{IEEEkeywords}

\begin{figure*}[htb]
\begin{center}
\begin{tabular}{c}
\includegraphics[width=\linewidth]{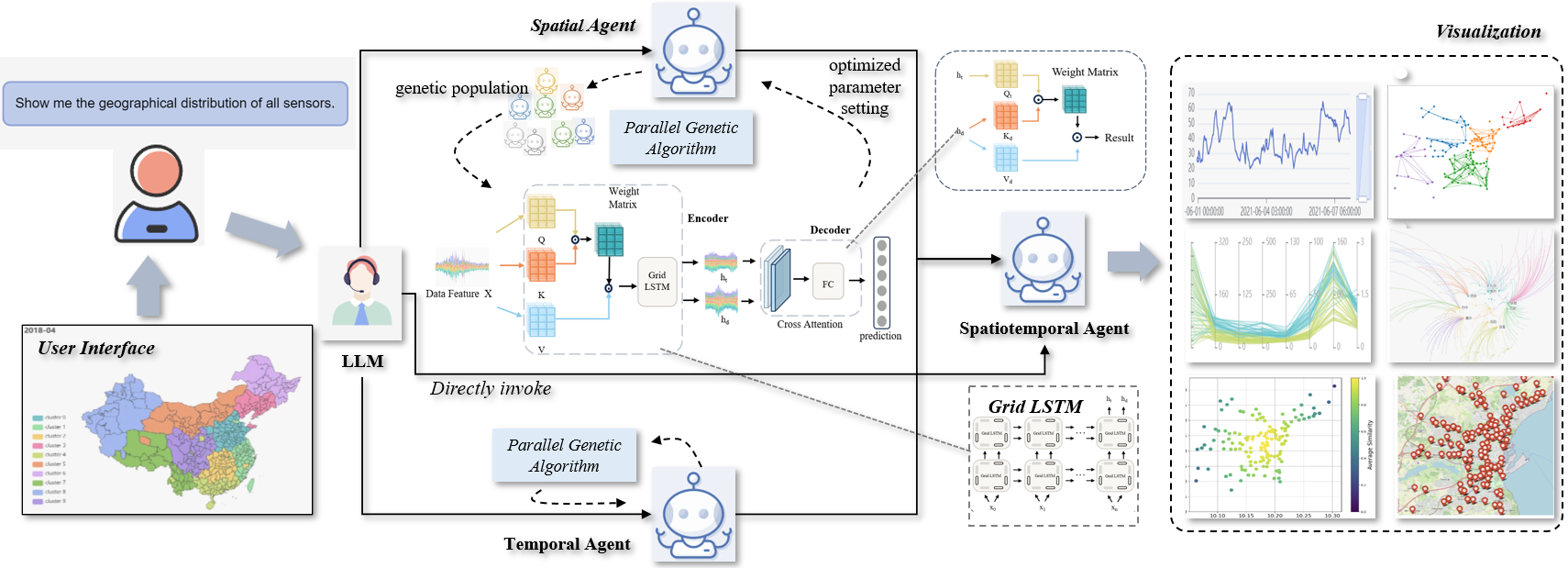} 
\end{tabular}
\vspace{-3mm}
\caption{User is using CityGPT to interact.}
\vspace{-3mm}
\label{fig:user_process}
\end{center}
\end{figure*}

\section{Introduction}
The concept of IoT was first introduced by Professor Kevin Ashton in the 1990s~\cite{GanZhixiang}. The IoT possesses the capability of automatic organization, information sharing, data, and resource management, enabling it to respond to and take action on environmental conditions and changes. In recent years, with the rapid development of emerging technologies such as big data and artificial intelligence, the scale and application of the IoT have expanded rapidly. Analyzing IoT data can help us better understand and grasp various situations and changes in the IoT environment. Through the analysis of large amounts of sensor data, real-world states and trends can be monitored and evaluated in real time, allowing for the timely identification of issues and corresponding actions~\cite{mirzai2023scheduling,khan2024float}. However, certain characteristics of IoT data pose challenges to its analysis, such as its large-scale volume, diverse dimensions, and complex interrelationships among these dimensions.


In the realm of extensive, diversified, and intricate IoT data, data visualization emerges as an indispensable analytical instrument. Its widespread application in IoT data analysis tasks, such as urban transportation~\cite{jin2021ecolens,feng2020topology, weng2020towards}, sensor networks~\cite{wei2021urban}, and meteorology~\cite{li2023visual}, underscores its pivotal role. Visualization serves to streamline the comprehension process by furnishing an intuitive portrayal of data, thereby augmenting the efficiency of knowledge generation~\cite{wang2016survey}. Nevertheless, when confronted with intricate datasets (e.g., high-dimensional data~\cite{xia2017ldsscanner}, graphical data~\cite{giovannangeli2020toward}, heterogeneous data~\cite{chen2018relationlines}, etc.) or engaged in sophisticated analysis endeavors (e.g., data prediction~\cite{riveiro2017anomaly}), supplementary aid for human interpretation remains imperative.

With the widespread adoption of deep learning, Convolutional Neural Networks (CNNs), Recurrent Neural Networks (RNNs), and Transformer models possess inherent advantages in spatiotemporal data analysis compared to traditional methods, owing to their hierarchical feature learning capabilities \cite{fukushima1980neocognitron, lipton2015critical, vaswani2017attention}. In contrast to conventional approaches, these deep learning models exhibit robust automatic feature extraction and function approximation abilities, enabling them to automatically analyze and extract features from spatiotemporal data and analyze their spatiotemporal information \cite{wang2020deep}. However, in the field of spatiotemporal data visualization, the challenge lies in the rational application of these data analysis models and the presentation of data information in effective, clear, and interactive visual forms. The latest advancements in Artificial Intelligence (AI) and Natural Language Processing (NLP) have ushered in a promising new era. Large-scale language models such as ChatGPT demonstrate remarkable common sense, reasoning, and planning capabilities\cite{ouyang2022training}. Natural language interfaces for visualization can generate visual results for users without the need for programming or technical construction, providing a flexible and intuitive means of data interaction.


In response to the above problems and challenges, we propose CityGPT, a spatial-temporal agent framework for IoT data learning and analysis with adaptive neural network structures. The system diagram is shown in Fig.~\ref{fig:user_process}. The primary contributions of this paper are as follows:

\begin{itemize}
\item \textbf{Deep Learning Analysis of Temporal Attributes in IoT Data.}
We propose a deep learning model that combines cross-attention mechanisms with GridLSTM networks to predict features specific to IoT data. Moreover, addressing the challenges of parameter tuning complexity and high user experience requirements associated with deep learning models, this study employs parallel genetic algorithms (PGA) to automatically optimize and configure model hyperparameters, thereby reducing the cost of model learning and enhancing the efficiency and speed of model deployment.

\item \textbf{Graph-Based Similarity Analysis of Spatial Attributes in IoT Data.}
To comprehensively exploit the spatial characteristics of IoT data, this paper enhances the CoSimHeat graph similarity calculation algorithm, extending its application from single-graph node computation to multi-graph node computation. By treating spatiotemporal data as a type of graph data, this algorithm calculates inter-regional similarity of geographical areas involved in IoT spatiotemporal data, effectively extracting spatial feature relationships between different regions.

\item \textbf{AI Agent System for Visualization of IoT Data Analysis.}
We propose an AI agent system that integrates a large language model. This system comprises four agents: text agent, temporal agent, spatial agent, and spatiotemporal fusion agent. By leveraging the large language model to parse user inputs into tasks comprehensible to the system and dispatching them to the corresponding agents, IoT data analysis and visualization are realized.
\end{itemize}

\section{Background and Motivation}
\subsection{IoT Time Series}
A time series is a sequence of numbers in chronological order. Time series data generated by massive sensors is becoming the most widespread in IoT computing. Time dependency is a very important feature for the IoT data, which has a great influence on sequential learning and analysis~\cite{WuTII2019,LinKDD2020,BoniolSIGMOD22}. For example, the data at the current time point is likely to be related to a previous time point or a time point in the long past. This is called short-term and long-term time dependency~\cite{AudibertDD2020}. It is necessary to design a feasible method to make good use of this feature for IoT applications, considering the traditional time series approaches have inefficiency in processing short-term dependency and long-term dependency at the same time.

\vspace{3pt} \noindent \textbf{\textit{IoT Time Series Features:}}
The IoT time series presents specific features:
\begin{itemize}
    \item \textbf{Large Scale}. Massive sensor data are gathered by distributed IoT devices. There are plenty of sensor data generated all the time.
    \item \textbf{Heterogeneous}. There are a variety of data acquisition devices so that the types of collected data are also different in the IoT.
    \item \textbf{High Dimensional}. Sensor data gathered by devices that are placed at geographical locations are labeled with time stamps. Data streams are measurements of continuous physical phenomenon. The spatial-temporal dynamics within data streams are inherent.
    \item \textbf{Time Sensitive}. The collected data should be processed with low latency to ensure their validity. For example, if a sensor device is running abnormally, it needs to be immediately detected to avoid affecting the normal operation of other devices.
\end{itemize}

\vspace{3pt} \noindent \textbf{\textit{Challenges:}} These IoT-specific features poses great challenges on \emph{accuracy}, \emph{reliability} and \emph{stability}. Therefore, a unified learning and analysis framework is expected to be deployed to address these challenges.

\subsection{AI Agent System}
The rapid development of AI agents has facilitated the transition of machine learning from the primary theoretical research stage to the practical application stage. This revolutionary advancement has facilitated the IoT system to evolve from a paradigm that uses mathematical tools to analyze time series data to a collaborative framework that integrates various functions. Recent advances in large-scale language models (LLMs) may provide an opportunity to help with interaction anxiety~\cite{qian2023,lin2023}. These generative models have the ability to effectively model a wide range of perspectives, roles, and expertise in a given domain.
Nowadays, LLMs have emerged as a potent text-to-task agent to empower systems; for instance, Zhang et al.~\cite{Zhangarxiv2023} integrated ChatGPT with a traffic prediction framework that enables users to use text and interact with traffic systems. Li et al.~\cite{LI2023nips} promote autonomous collaboration between exchange agents and delve into their "cognitive" processes, in order to study cooperative behavior and the capabilities of multi-agent systems.

\vspace{3pt} \noindent \textbf{\textit{Challenges:}} However, in the realm of IoT, the biggest problem
is the huge amount of data and the complex structure of time series data, which makes it difficult to agentize in a generalized way for systems. Combining high performance and generalizability for specific tasks seems elusive.

\subsection{Opportunities of AI Agent System for IoT} 
\vspace{3pt} \noindent  Due to the constraints of limited bandwidth and computation resources, many IoT applications need to preload the IoT data. 

\vspace{3pt} \noindent \textbf{\textit{Motivation for CityGPT:}} There are application and interaction needs for designing AI agents for IoT learning and analysis.

\textit{Application Needs:} The agile platform is expected to integrate the capabilities of connectivity, storage, computing and analysis as an end-to-end solution~\cite{YangSenSys2021,DengSenSys2022}. The paradigm of end-to-end learning emerges to meet such requirements~\cite{ParkSIGMOD2022}. It is a technique in the context of AI and ML where the model learns all the steps between the initial input phase and the final output result. This is a deep learning process where all of the different parts are simultaneously trained instead of sequentially. For most deep learning models, the complexity of model construction, unreasonable model hyperparameter settings, and long model training process bring great obstacles to the training process, which greatly affects the promotion and application of the learning model. 

\textit{Model Interaction Needs:} Introducing complex models increases the complexity of interaction, which is unfriendly to the wider use of the system, narrowing the range of applications. For a system built for IoT consideration, we need to consider the interactivity of functions inside the system. Agentizing functions as separated AI for better functional clarity and introducing text interaction to optimize the system.

\vspace{3pt} \noindent \textbf{\textit{Challenges and Solutions:}} In our paper, we employ AutoML as one of the agents in CityGPT to tackle time series challenges. Hyperparameter optimization (HPO) is crucial in model training and significantly impacts model performance. As part of AutoML, HPO is integrated into the training, validation, and testing phases to provide adaptive capabilities within the unified framework for Learning and Analysis of IoT time series. Additionally, our system includes temporal, spatial, and spatiotemporal agents to enhance interactivity and broaden application scope.


\section{System Overview}
\begin{figure}[htb]
\begin{center}
\begin{tabular}{c}
\includegraphics[width=\linewidth]{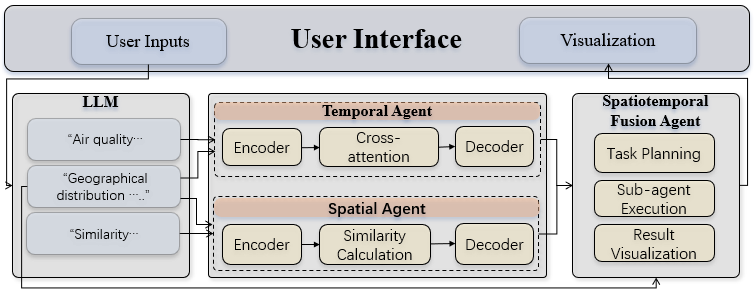} 
\end{tabular}
\vspace{-3mm}
\caption{System overview.}
\vspace{-3mm}
\label{fig:sys}
\end{center}
\end{figure}

The IoT data analysis framework proposed in this paper is depicted in Fig.~\ref{fig:sys}. The upper section illustrates the user interaction interface, encompassing user requirement input and visualization functionality for displaying results. The lower section delineates the primary components of the system: the requirement agent, which translates user inputs in natural language into executable tasks for the system; the analysis agent, comprising temporal and spatial agents, enabling analysis of IoT data from both temporal and spatial dimensions; and the spatiotemporal fusion agent, responsible for visualizing the results of data analysis tasks parsed by the requirement agent.

\noindent \textbf{Requirement Agent.} The requirement agent allows users to input demands in natural language by integrating a large language model (such as GPT). This agent parses textual inputs into tasks that can be executed by various system agents and distributes these tasks.

\noindent \textbf{Analysis Agent.} The analysis agent is divided into temporal and spatial agents. The temporal agent utilizes a GridLSTM network model incorporating cross-attention mechanisms to predict and analyze features of IoT data. To enhance model training efficiency and reduce task execution time, this study employs parallel genetic algorithms to optimize and accelerate the configuration of model hyperparameters, thereby decreasing model configuration time. The spatial analysis agent utilizes an extended CoSimHeat to compute spatial similarities of IoT data and employs clustering analysis to extract spatial information from the data.

\noindent \textbf{Spatiotemporal Fusion Agent.} The spatiotemporal fusion agent is responsible for presenting the results of user-inputted task analyses to users in visual form. This agent receives analysis result data from the temporal and spatial analysis agents and, by invoking the corresponding sub-visualization agents, showcases the system's final analysis results.

\section{Design}
\subsection{Temporal Agent}
\subsubsection{Prediction Model Design}
\begin{figure}[tb]
	\centering
	\includegraphics[width=\linewidth]{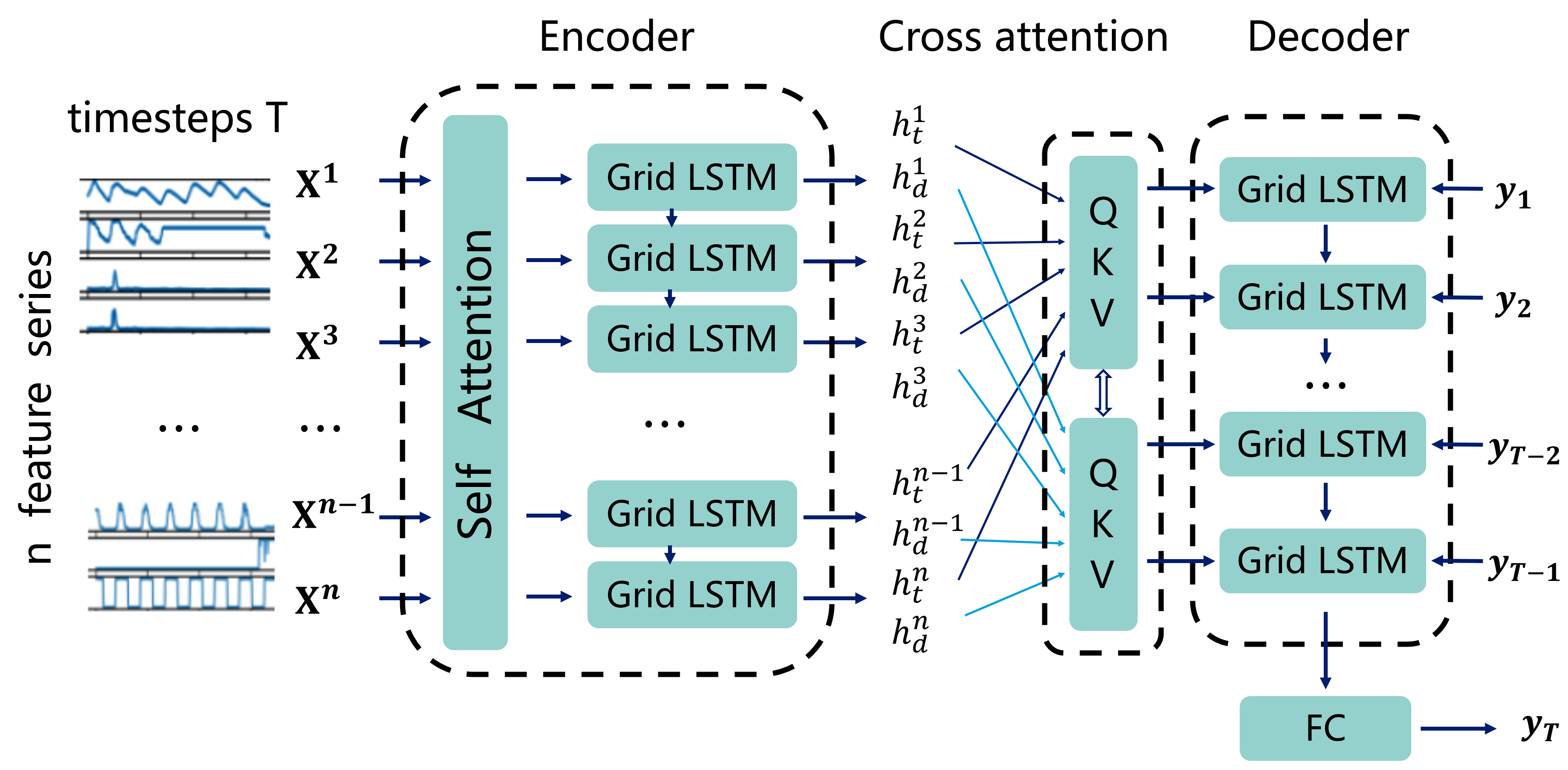} 
        \vspace{-3mm}
 \caption{Attentional Grid LSTM model for prediction on IoT time series.} 
        \vspace{-3mm}
	\label{fig:gridlstm-attn} 
\end{figure}

In order to predict the characteristics of spatiotemporal data in IoT, this paper proposes a GridLSTM network model based on the cross-attention mechanism to analyze short-term and long-term time dependency. The Attentional Grid LSTM consists of two stages: {Encoder Stage} and {Decoder Stage}, as illustrated in Fig.~\ref{fig:gridlstm-attn}.

\begin{figure}[tb]
\begin{center}
\begin{tabular}{cc}
\includegraphics[width=3.3cm]{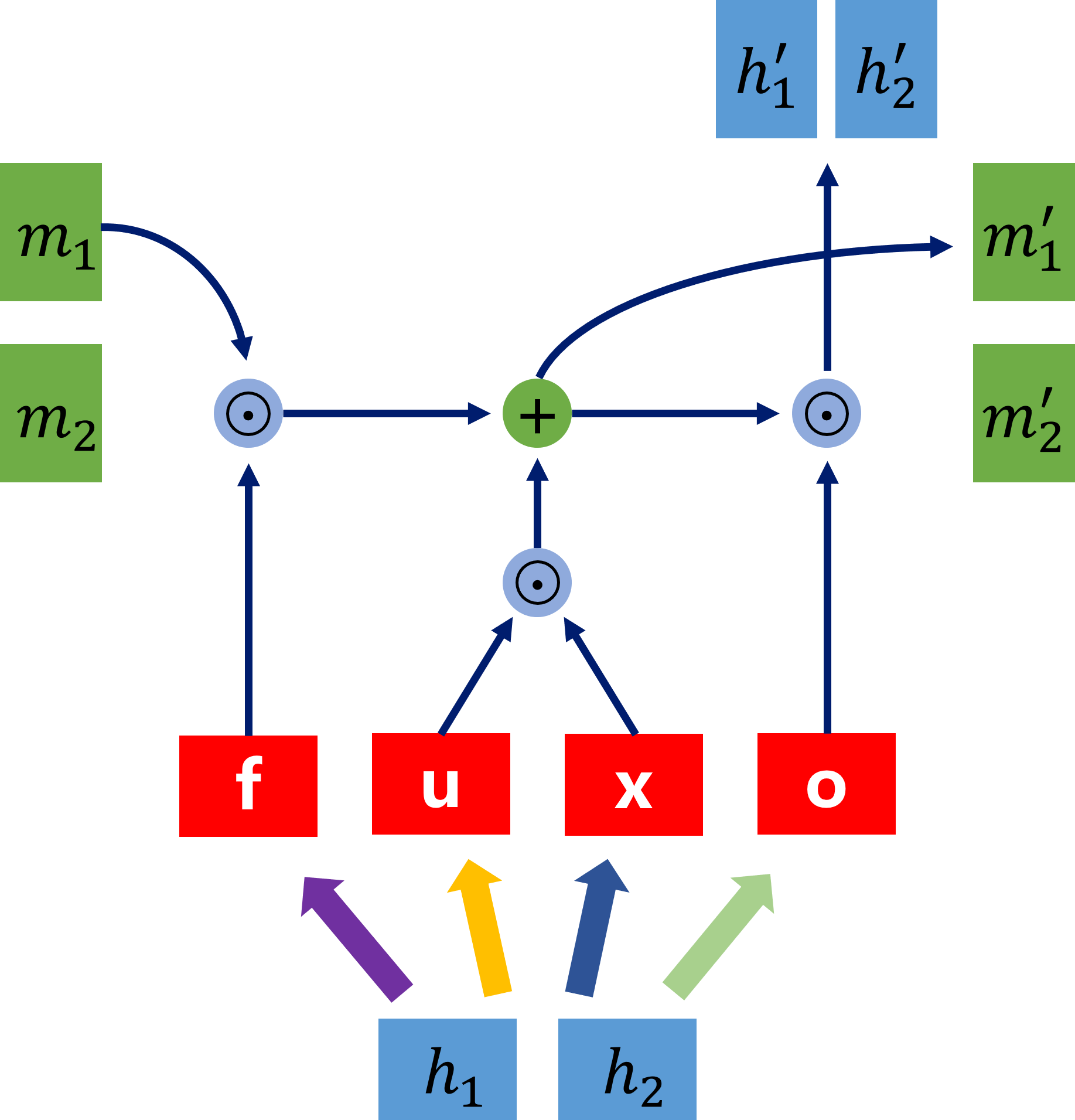} & \includegraphics[width=4.8cm]{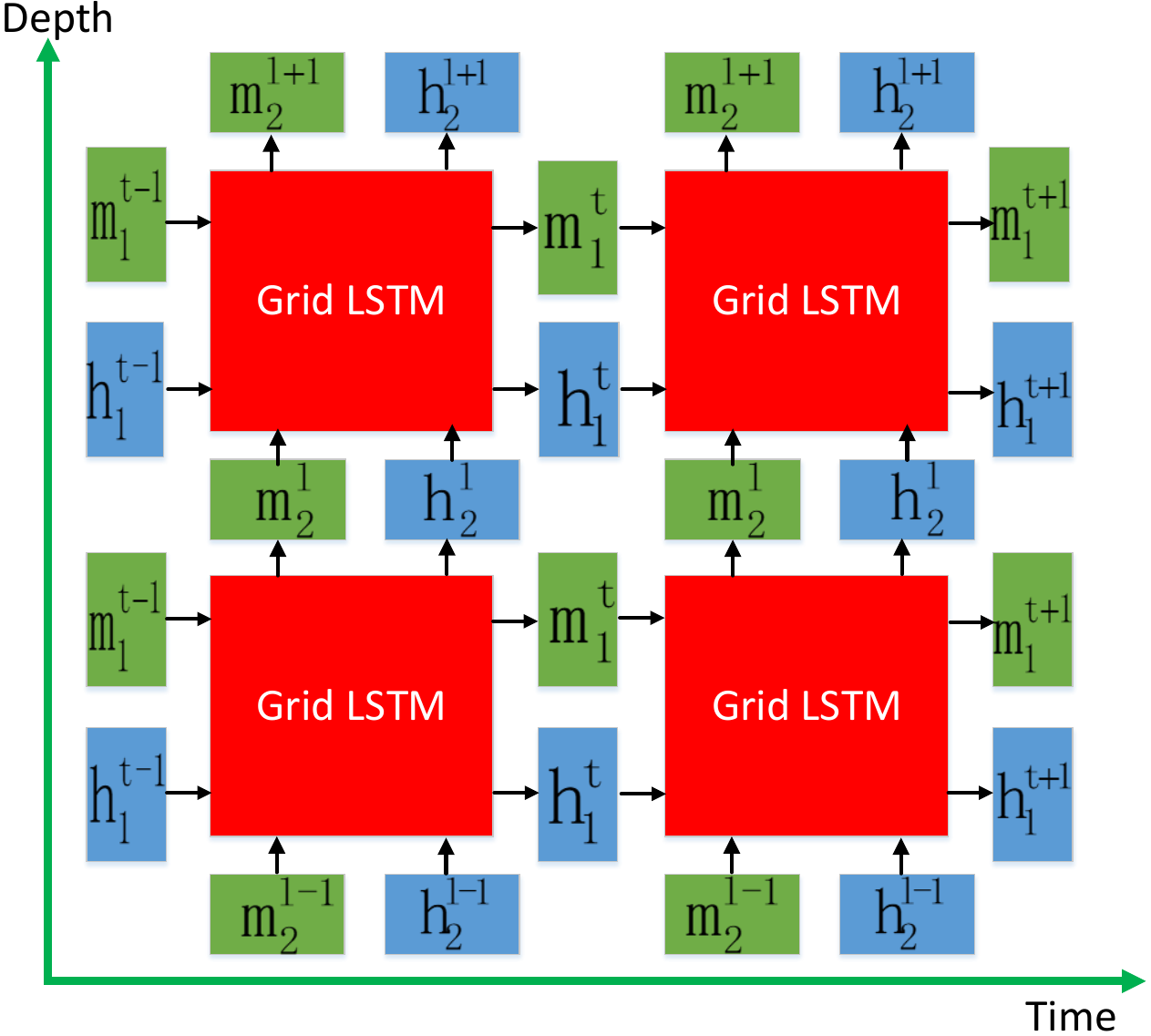} \\
{\scriptsize(a) Grid LSTM Cell} & {\scriptsize (b) Grid LSTM Network}
\end{tabular}
\vspace{-3mm}
\caption{Grid LSTM cell and network structure.}
\vspace{-3mm}
\label{fig:LSTM}
\end{center}
\end{figure}

As shown in Fig.~\ref{fig:LSTM}, it is a Grid LSTM network composed of four cells fully connected by recurrent connections, with the horizontal axis representing the time dimension and the vertical axis representing the depth dimension. Each cell of the lower hidden layer in the network is connected to the cell in the higher hidden layer through the feed-forward connection in both time and depth dimension. Only the current state and some of the previous states are needed to train the network.

Specifically, the input in the time dimension corresponds to a time series, and each hidden layer's cell corresponds to a time step of the time series. For the output of the current time step, it takes the data from different time steps into account and evaluates their impact at the next time step. For each Grid LSTM cell, it controls the input, storage and output of data by the gate mechanisms. The gate units of current cell receive the output of previous time step generated at previous hidden layer (such as $m_1^{t-1}$) and the input of current time step sample, and then determine the output of current hidden layer (such as $m_1^{t}$), which will be used at the next time step to generate the output of next hidden layer (such as $m_1^{t+1}$). As for the depth information process along the vertical dimension, it follows the similar workflow as the time dimension; the gate units of current cell process the time series from previous hidden layer (such as $m_2^{l-1}$) and generate the output of current hidden layer (such as $m_2^{l}$); the output of next hidden layer (such as $m_2^{l+1}$) is derived accordingly.

To improve the prediction performance of IoT time series data, our CityGPT framework combines the Grid LSTM network with an attention mechanism, which assigns different weights to different feature series in the input layer and selects the most relevant features according to the weights. The details of each stage are described below: 

\vspace{5pt} 
\noindent\textbf{Encoder Stage.}
Let $x$ be a time series of length $T$:

\begin{small}
\begin{equation}
\label{equ:encoder1}	
x=\{x_1, x_2, \cdots, x_T\}, \textrm{ with } x_i=(x_i^1, x_i^2, \cdots, x_i^N)^{\top} \in \mathbb{R}^N
\end{equation}
\end{small}

\noindent
where each $x_i \ (i=1, 2, \cdots, T)$ has $N$ features (dimensions), and $x_i^j$ corresponds to the $j$-th feature of $x$ at time $i$.

For each time series $x$, we first define \emph{the attention weight} of $x$ as

\begin{small}
\begin{equation}
\label{equ:encoder2}	
\alpha=\{\alpha_1, \alpha_2, \cdots, \alpha_T\}, \textrm{ with } \alpha_i=(\alpha_i^1, \alpha_i^2, \cdots, \alpha_i^N)^{\top} \in \mathbb{R}^N
\end{equation}
\end{small}

\noindent
where each $\alpha_i^j$ denotes the weight of the $j$-th feature at time~$i$.

Then, integrating the attention weight $\alpha$ to the time series $x$, we generate new weighted time series $\hat{x}$ as follows:

\begin{small}
\begin{equation}
\label{equ:encoder3}	
\hat{x}=\{\hat{x}_1, \cdots, \hat{x}_T\}, \textrm{ with } \hat{x}_i=(\alpha_i^1 x_i^1, \cdots, \alpha_i^N x_i^N)^{\top} \in \mathbb{R}^N
\end{equation}
\end{small}

Finally, we obtain the hidden state $h_t$ at time $t$ and the hidden state $h_d$ in depth.
This step can be rewritten as a function:

\begin{small}
\begin{equation}
\label{equ:encoder4}
h_{t}^{i},  h_{d}^{i} = \textsf{GridLSTM}({\hat{x}_{i}}) \quad (\forall i=1,2,\cdots,T)
\end{equation}
\end{small}

\vspace{5pt}
\noindent\textbf{Decoder Stage.}
We leverage Grid LSTM based on the cross attention model to decode the output information from the encoder. We use the encoder outputs $h_{t}$ and $h_{d}$ as inputs to the cross attention module to extract important information from the hidden states. The extracted information is then fed into the same Grid LSTM network for decoding. Finally, the decoded information is fed to the fully connected network layer and then the final output $\hat{y}_{T}$ is obtained.

\subsubsection{Hyperparameter Optimization using PGA}
\label{sect:Method}

The PGA algorithm presented in this paper runs in three steps: initialization, iteration, and termination. The specific execution flow is depicted with algorithm explanation in Algorithm~\ref{algo:01}.
In the initialization phase (lines~\ref{ln:a01-01}--\ref{ln:a01-02}) in Algorithm~\ref{algo:01}, we will divide the population and encode individuals. The algorithm will first divide the populations equally according to the set numbers. Since the hyperparameters involved in this paper are all integers, we use binary-encoded individuals. The hyperparameters to be optimized are stored as individuals in subpopulations by a fixed-length string of binary numbers.

\vspace{-2mm}
\begin{algorithm}[h]
\small
\DontPrintSemicolon
\SetKwInOut{Input}{Input} 
\SetKwInOut{Output}{Output} 
\Input{Population size $m$, Number of subpopulations $N$, Number of algorithm iterations $K_{1}$, Number of subpopulation iterations $K_{2}$, Fitness function $F$}
\Output{Global optimal solution $Y^{by}$}
\nl \label{ln:a01-01} generate initial population of $m$ chromosomes $y^{i}$, $(i = 1,2,\ldots,m)$ \;
\nl \label{ln:a01-02} delineate subpopulations, the number of individuals in the subpopulation $n=\frac{m}{N}$ \;
\nl \label{ln:a01-03} \For{$k_{1}:=1, 2, \cdots, K_{1}$}{
\nl \label{ln:a01-04} \For{$k_{2}:=1, 2, \cdots, K_{2}$}{
\nl \label{ln:a01-05} calculate the value of the fitness function for all individuals $f^{i}_{k_{2}}= F(y^{i})$, $(i = 1,2,\ldots,n)$\;
\nl \label{ln:a01-06} select individuals $O$ based on $f^{i}_{k_{2}}$\;
\nl \label{ln:a01-07} complete crossover among $O$\;
\nl \label{ln:a01-08} complete variation among $O$, generate new individuals $O'$\;
\nl \label{ln:a01-09} $y^{i} \leftarrow O'$\;
}
\nl \label{ln:a01-10} perform inter-subpopulation exchange\;
}
\nl \label{ln:a01-11} calculate the value of the fitness function of the individual after the iteration $f^{i}=F(y^{i})$\;
\nl \label{ln:a01-12} select the optimal individual according to $f'$, $F(Y^{by})=\min\{f^{i}\}$\;
\nl \label{ln:a01-13} \Return $Y^{by}$\;
\caption{Parallel Genetic Algorithm}
\label{algo:01}
\end{algorithm}
\vspace{-2mm}

In the iterative phase (lines~\ref{ln:a01-03}--\ref{ln:a01-10}), we will set up the fitness function, selection techniques, crossover operators, mutation operators, and subgroup exchange to achieve the iterative solution of the algorithm. The fitness function reflects the degree of superiority or inferiority of the solution corresponding to this individual. Therefore, we take the prediction error $RMSE$ of the deep learning model corresponding to the hyperparameter value of this individual as the fitness function of this individual (lines~\ref{ln:a01-05}). Selection techniques refer to the selection of good individuals from the population and the elimination of inferior individuals (lines~\ref{ln:a01-06}). We choose the tournament selection technique~\cite{katoch2021review}. This approach randomly selects a subset of individuals from the subpopulation and then identifies the best individuals based on their fitness function.

Following selection, individuals undergo uniform crossover and lip bit mutation (lines~\ref{ln:a01-07}--\ref{ln:a01-08}). Subsequently, new individuals are generated within each subpopulation, with the least fit individuals removed (line~\ref{ln:a01-09}). This process continues iteratively until the specified number of generations is reached, at which point subpopulation exchange occurs (line~\ref{ln:a01-10}). We employ a sub-group migration method proposed by Arkhipov et al.~\cite{arkhipov2015simple} to facilitate this exchange. Initially, individuals are selected based on fitness results and exchanged between processes. Then, individuals selected in the previous process are passed to the next, and the current process receives individuals from the previous one. Finally, best-fit individuals from the current subgroup form a new subgroup. This iterative process optimizes hyperparameters in the PGA (lines~\ref{ln:a01-11}--\ref{ln:a01-12}).

After the number of iterations reaches the set number, the most suitable individual among all subpopulations is selected, and the iteration is terminated. The hyperparameters represented by this individual are decoded and set as the final optimization result of the PGA as the hyperparameters of the deep learning model.

\subsection{Spatial Agent}
\label{sect:similarityModel}
\begin{figure}[tb]
\begin{center}
\begin{tabular}{cc}
\includegraphics[width=0.47\linewidth]{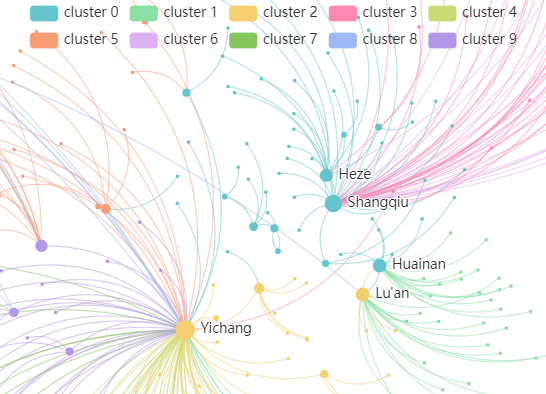} & 
\includegraphics[width=0.47\linewidth]{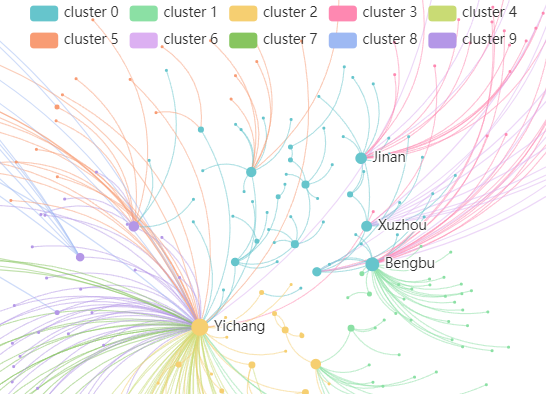} \\
{\scriptsize(a) Spatial similarity of April} & {\scriptsize (b) Spatial similarity of June}
\end{tabular}
\vspace{-3mm}
\caption{Example of spatial similarity.}
\vspace{-3mm}
\label{fig:motiExample}
\end{center}
\end{figure}

\textbf{Necessity of Spatial Agent.} We initially used a similarity algorithm to cluster the cities using a pollutant-related dataset, in order to ensure its effectiveness in estimating spatial similarity. Similarity results of the April and June 2018 are depicted in Fig.~\ref{fig:motiExample}. In the figure, each dot represents a city, and the number of lines connected to each dot represents the number of other cities that are similar to that city, which reflects the city's level of importance in the network. In this graph, the names of cities connected to more than 15 cities are labeled. By looking at the changes in the graph, we can see that the key cities in the overall city network have changed over time. In April, the key cities include Yichang, Bengbu, Xuzhou, and Jinan. However, in June, all key cities except Yichang have changed. This suggests that the structural center of gravity of the entire city network has changed over time. This change is reflected in the distribution of pollutant data, showing that there are spatial differences in the distribution characteristics of pollutants at the same point in time. This reveals the necessity of using a similarity-based agent to capture spatial features in our framework.

\textbf{CoSimHeat.} In the era of Big Data, countless web applications require effective similarity measures based on billion-scale network structures. To efficiently capture spatial features, it is imperative to utilize a promising graph-theoretic similarity model that iteratively captures the notion that "two different nodes are evaluated as similar if they are connected to similar nodes.". In our framework, to effectively process spatial features of IoT data, we use CoSimHeat, which focuses on the structure information of graphs, to calculate the similarity. Compared with CoSimRank~\cite{rothe2014acl},  which depends on the existence of directly reachable edges at two nodes, CoSimHeat performance is better in IoT-related scenarios. 

CoSimHeat is a heat kernel graph diffusion model used to compute the similarity of individual nodes within a single graph~\cite{yu2022cosimheat}. Building upon this, our paper extends its applicability from a single graph to two graphs, enabling cross-graph node computation. We apply both the original CoSimHeat and the extended similarity calculation algorithm to the task of spatiotemporal data spatial feature analysis.


Define two graphs, $G_1$ and $G_2$, with nodes $ x $ and $ y $ respectively. For node $ x $ in $ G_1 $, $ O_x $ and $ I_x $ denote its out-neighbour and in-neighbour sets, while $ |O_x| $ and $ |I_x| $ represent its respective out-degree and in-degree. Similarly, for node $ y $ in $ G_2 $, $ O_y $ and $ I_y $ represent its out-neighbour and in-neighbour sets, and $ |O_y| $ and $ |I_y| $ denote its out-degree and in-degree. Let $ T_{x,y}(t) $ denote the average temperature of nodes $ x $ and $ y $ at time $ t $. Now, during the time interval $ [t, t+\Delta t] $, consider the heat diffusion from nodes $ x $ and $ y $ to their neighboring nodes. The heat generated by $ x $ and $ y $ is denoted as $ H^+(T_{x,y},t,\Delta t) $, while the heat received from other nodes is denoted as $H^-(T_{x,y},t,\Delta t)$. Assuming equal heat diffusion capability for each node to its neighbors, the heat change for the node pair $ (x, y) $ during $ \Delta t $ depends on three components: (1) the heat received by node $ x $ from its neighbors, (2) the heat received by node $y$ from its neighbors, and (3) the heat diffusion between the node pair $(x, y)$. Based on this, the equation for the heat change of the node pair at time $\Delta t$ can be formulated as follows:

\begin{small} 
\begin{equation}
\label{equ:Cosim1}
    \begin{split}
    &T_{x,y}(t+\Delta t) - T_{x, y}(t) = \\ 
    &\sum_{x'\in{I_x}} H^-_{x'}(T_{x',y},t,\Delta t) +
    \sum_{y'\in{I_y}} H^-_{y'}(T_{x,y'},t,\Delta t) -
    H^+(T_{x,y},t,\Delta t)
    \end{split}
\end{equation}
\end{small}

Regarding the last term of Eq.~\eqref{equ:Cosim1}, it can be simplified to Eq.~\eqref{equ:Cosim2} by introducing a damping factor $\lambda$. This implies that the heat diffusion of node pairs within the time interval $t$ is directly proportional to the temperature $T_{x,y}$ of the node pair and the diffusion time. This proportionality factor $\lambda$ is defined to characterize the degree of difficulty of heat diffusion.

\begin{small}
\begin{equation}
\label{equ:Cosim2}
    H^+(T_{x,y},t,\Delta t) = \lambda T_{x,y}(t)\cdot \Delta t 
\end{equation}  
\end{small}

Based on the preceding assumption that each node distributes heat equally to its neighboring nodes, it follows that the neighboring nodes of nodes $x$ and $y$ will evenly share the heat propagated by the node pair $(x, y)$. Conversely, the heat received by nodes $x$ and $y$ from their neighboring nodes during the time interval $\Delta t$ can be expressed as Eq.~\eqref{equ:Cosim3} and Eq.~\eqref{equ:Cosim4}:

\begin{small}
\begin{equation}
\label{equ:Cosim3}
    H^-(T_{x',y},t,\Delta t) = \frac{\lambda}{2|O_{x'}|} T_{x',y}(t)\cdot \Delta t, \qquad  x'\in I_x 
\end{equation}
\end{small}

\begin{small}
\begin{equation}
\label{equ:Cosim4}
    H^-(T_{x,y'},t,\Delta t) = \frac{\lambda}{2|O_{y'}|} T_{x,y'}(t)\cdot \Delta t,  \qquad y'\in I_y 
\end{equation}
\end{small}

Substituting Eqs.~\eqref{equ:Cosim2}--\eqref{equ:Cosim4} into Eq.~\eqref{equ:Cosim1}, the expression for the heat change of the node pair $(x, y)$ becomes as shown in Eq.~\eqref{equ:Cosim5}.

\begin{small}
\begin{equation}
\label{equ:Cosim5}
    T_{x,y}(t+\Delta t) - T_{x, y}(t) = \frac{\lambda \cdot \Delta t}{2} (
    \sum_{x'\in{I_x}} \frac{T_{x',y}(t)}{|O_{x'}|} + \sum_{y'\in{I_y}} \frac{T_{x,y'}(t)}{|O_{y'}|} - 2T_{x,y}(t))
\end{equation}
\end{small}

According to calculus principles, as ${\Delta t \to 0}$, Eq.~\eqref{equ:Cosim5} becomes a differential equation with respect to the variable $T_{x, y}$ , as shown below:

\begin{small}
\begin{equation}
\label{equ:Cosim6}
    \frac{{\rm d} T_{x,y}(t)}{{\rm d} t} = \frac{\lambda}{2} (M_{x,y}(t)+N_{x,y}(t)-2T_{x,y}(t))
\end{equation}
\end{small}

\begin{small}
\begin{equation}
\label{equ:Cosim7}
    M_{x,y}(t) = \sum_{x'\in{I_x}} \frac{T_{x',y}(t)}{|O_{x'}|}, 
    N_{x,y}(t) = \sum_{y'\in{I_y}} \frac{T_{x,y'}(t)}{|O_{y'}|}
\end{equation}
\end{small}

Thus, the heat diffusion differential equation for the node pair $(x, y)$ has been derived. By integrating the variations in node similarity with this differential equation, the similarity matrix among nodes within the graph can be deduced. Given that this differential equation is a first-order ordinary differential equation with multiple solutions, to ensure solution uniqueness, we define the solution of this differential equation at time $t = 0$ as illustrated in Eq.~\eqref{equ:Cosim8}.

\begin{small}
\begin{equation}
\label{equ:Cosim8}
    T_{x,y}(0) = \frac{\theta}{m_1 \cdot m_2}(|O_x| \cdot |O_y|) + 
    \frac{1-\theta}{n_{S_0}} \cdot 
    \left\{
        \begin{aligned}
            & 1, ((x, y) \in S_0)\\
            & 0, ((x, y) \notin S_0)
        \end{aligned}
    \right.
\end{equation}
\end{small}

\noindent
where $m_1$ and $m_2$ represent the number of edges in graphs $G_1$ and $G_2$, respectively. $S_0$ denotes the set of node pairs in the two graphs known to have relationships, with $n$ representing the number of node pairs in this set. The hyperparameter $\theta$, ranging from 0 to 1, is a weighting factor used to balance the influence of the first and second terms in the equation.In the experiments of this paper, we set $\theta=0.5$.


\begin{figure*}[tb]
\begin{center}
\begin{tabular}{cc}
\includegraphics[width=0.4\linewidth]{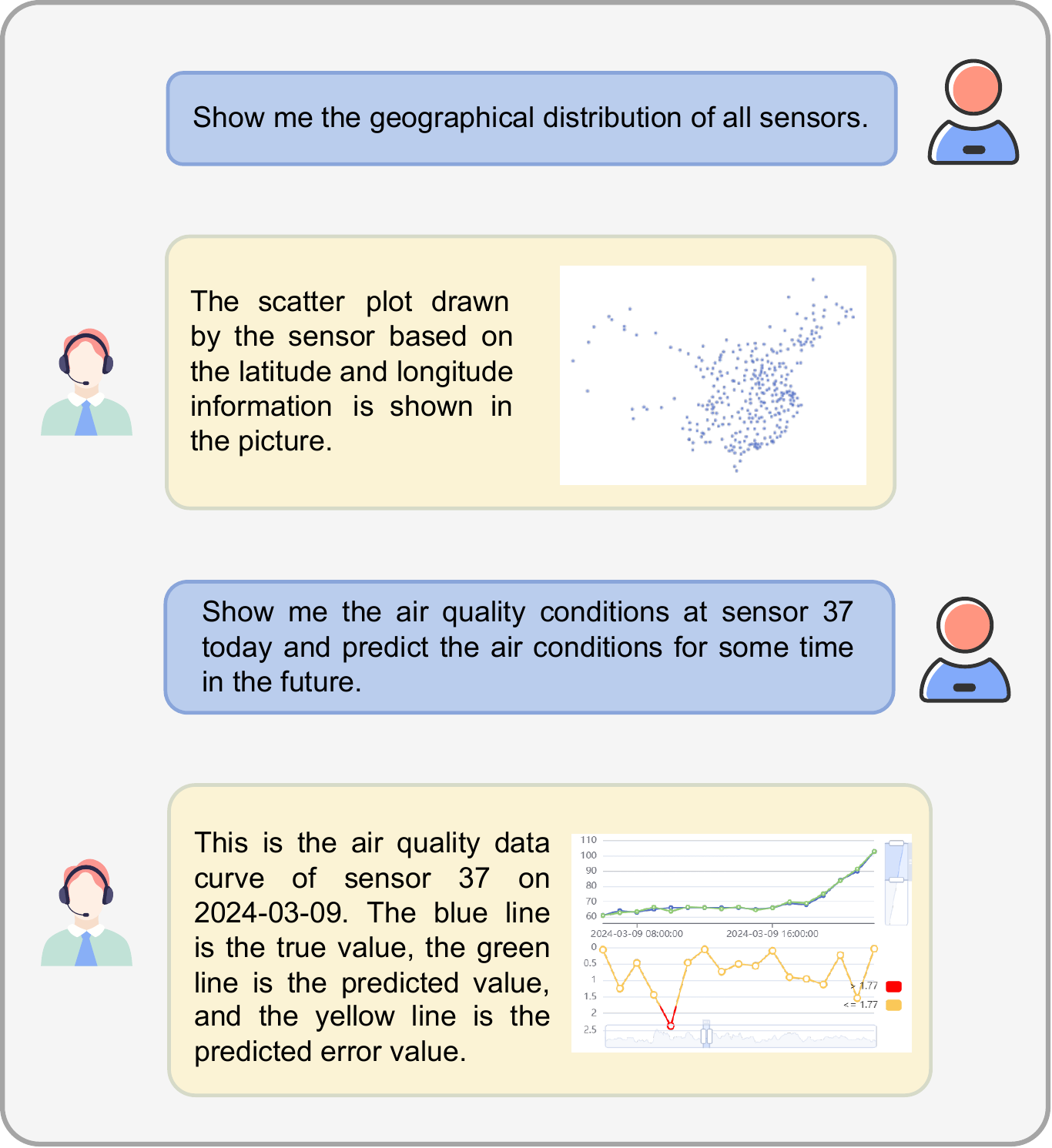} & 
\includegraphics[width=0.4\linewidth]{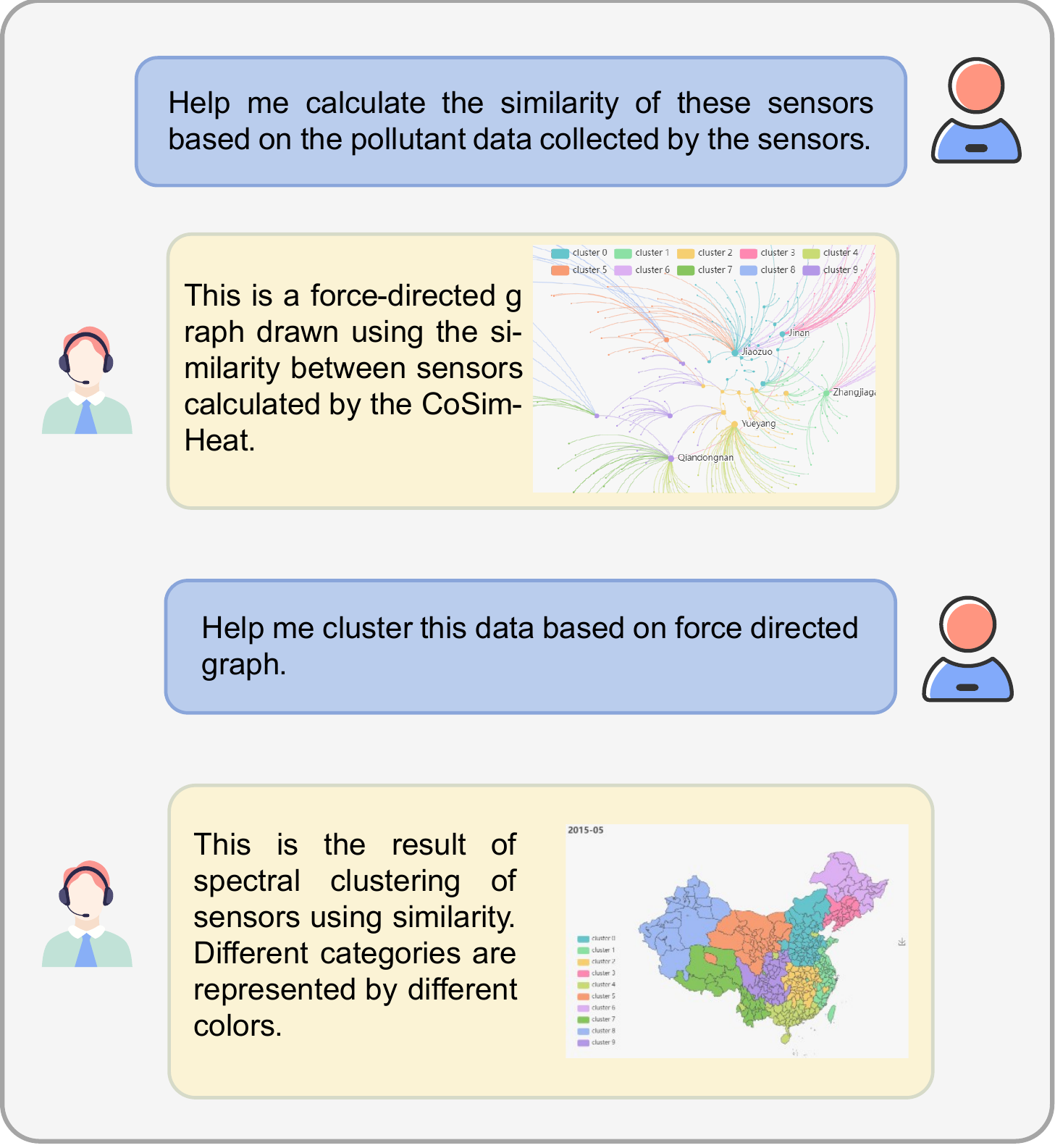} \\
{\scriptsize(a) Conversation 1} & {\scriptsize (b) Conversation 2}
\end{tabular}
\vspace{-3mm}
\caption{Example of spatiotemporal fusion agent task}
\vspace{-3mm}
\label{fig:taskExample}
\end{center}
\end{figure*}

The first term of the equation signifies the similarity of features between nodes, indicating that nodes with similar features tend to be alike. Specifically, it is assumed that nodes with higher out-degree centrality are more likely to exhibit similarity. The second term represents the "distance" similarity, implying that nodes requiring fewer intermediate nodes to connect are more similar. At the initial time step, the most similar node pairs can be directly obtained from $S_0$ containing known inter-graph node relationships.


Substituting  $|O_{x'}|=[diag(\boldsymbol{d}^+)^{-1}]_{x',x'}$ into Eq.~\eqref{equ:Cosim7}, we obtain Eq.~\eqref{equ:Cosim9}, where $A$ is the adjacency matrix of graph $G_1$, $diag(\boldsymbol{d}_1^+)$ is the out-degree matrix of graph $G_1$, and $T_t$ represents the similarity matrix composed of nodes from graphs $G_1$ and $G_2$.

\begin{small}
\begin{equation}
\label{equ:Cosim9}
    \begin{split}
    M_{x,y}(t) & = \sum_{x'\in{I_x}} \frac{T_{x',y}(t)}{|O_{x'}|}\\ 
    & = \sum\nolimits_{i = 1}^{m_1} [\boldsymbol{A}]_{i,y} \cdot [diag(\boldsymbol{d}_1^+)^{-1}]_{i,i} \cdot [\boldsymbol{T}_t]_{i,y} \\
    & = [\boldsymbol{A}^T \cdot diag(\boldsymbol{d}_1^+)^{-1} \cdot \boldsymbol{T}_t]_{x, y}
    \end{split}
\end{equation}
\end{small}

Similarly, the calculation process of $N_{x,y}(t)$ can be expressed as Eq.~\eqref{equ:Cosim10}, where $\boldsymbol{B}$ is the adjacency matrix of graph $G_2$, and $diag(\boldsymbol{d}_2^+)$ is the out-degree matrix of graph $G_2$.

\begin{small}
\begin{equation}
\label{equ:Cosim10}
    \begin{split}
    N_{x,y}(t) & = \sum_{y'\in{I_y}} \frac{T_{x,y'}(t)}{|O_{y'}|}\\ 
    & = \sum\nolimits_{j = 1}^{m_2} [\boldsymbol{B}]_{x,j} \cdot [diag(\boldsymbol{d}_2^+)^{-1}]_{j,j} \cdot [\boldsymbol{T}_t]_{x,j} \\
    & = [\boldsymbol{T}_t \cdot diag(\boldsymbol{d}_2^+)^{-1} \cdot \boldsymbol{B}]_{x, y}
    \end{split}
\end{equation}
\end{small}

Substituting Eq.~\eqref{equ:Cosim9} and Eq.~\eqref{equ:Cosim10} into Eq.~\eqref{equ:Cosim6}, we obtain Eq.~\eqref{equ:Cosim11} and Eq.~\eqref{equ:Cosim12}.

\begin{small}
\begin{equation}
\label{equ:Cosim11}
    \frac{{\rm d} \boldsymbol{T}_t}{{\rm d} t} = 
    \frac{\lambda}{2}(
    \boldsymbol{A}^T \cdot diag(\boldsymbol{d}_1^+)^{-1} \cdot \boldsymbol{T}_t +
    \boldsymbol{T}_t \cdot diag(\boldsymbol{d}_2^+)^{-1} \cdot \boldsymbol{B} - 2 \boldsymbol{T}_t
    )
\end{equation}
\end{small}

\begin{small}
\begin{equation}
\label{equ:Cosim12}
    \boldsymbol{T}_0 = \frac{\theta}{m_1 \cdot m_2}
    \boldsymbol{d}_1^+(\boldsymbol{d}_2^+)^T + 
    \frac{1-\theta}{n_{S_0}}S_0
\end{equation}
\end{small}

Then, to simplify Eq.~\eqref{equ:Cosim11}, we define Eq.~\eqref{equ:Cosim_P} and Eq.~\eqref{equ:Cosim_Q}.

\begin{small}
\begin{equation}
\label{equ:Cosim_P}
    \boldsymbol{Q} = \frac{\lambda}{2}[\boldsymbol{A}^T diag(\boldsymbol{d}_1^+)^{-1} - \boldsymbol{I}_A]
\end{equation}
\end{small}

\begin{small}
\begin{equation}
\label{equ:Cosim_Q}
    \boldsymbol{P} = \frac{\lambda}{2}[diag(\boldsymbol{d}_2^+)^{-1} \boldsymbol{B} - \boldsymbol{I}_B]
\end{equation}
\end{small}

\noindent
where $\boldsymbol{I}_A$ and $\boldsymbol{I}_B$ are the identity matrices of $\boldsymbol{A}^T$ and $\boldsymbol{B}$. 
The resulting differential equation is shown in Eq.~\eqref{equ:Cosim13}.


\begin{small}
\begin{equation}
\label{equ:Cosim13}
    \frac{{\rm d} \boldsymbol{T}_t}{{\rm d} t} = \boldsymbol{Q} \boldsymbol{T}_t + \boldsymbol{T}_t \boldsymbol{P}
\end{equation}
\end{small}

Furthermore, by combining with the knowledge of differential equations, the solution expression for the similarity matrix can be obtained as Eq.~\eqref{equ:Cosim14}. Thus, based on the adjacency matrices, out-degree matrices, edge counts of graphs $G_1$ and $G_2$, as well as the initial relationships between nodes in the two graphs, the similarity between any pair of nodes in these graphs can be calculated. This similarity matrix can then be constructed for analyzing the spatial distribution similarity across graphs.

\begin{small}
\begin{equation}
\label{equ:Cosim14}
    \boldsymbol{S}_0 = e^{\boldsymbol{Q}} \cdot \boldsymbol{T}_0 \cdot e^{\boldsymbol{P}} \quad \text{ where }
    e^{\boldsymbol{Q}} = \sum_{n=0}^{\infty} \frac{\boldsymbol{Q}^n}{n!}
\end{equation}
\end{small}

\subsection{Spatiotemporal Fusion Agent}
By integrating a large language model, the spatiotemporal fusion agent can decompose IoT data analysis tasks based on user input and present the results of temporal and spatial agents in a visual format to help users understand and analyze IoT data. The execution process of this agent generally consists of the following key steps:

\begin{enumerate}[{(1)}]
\item Task planning: Accepting task distribution from the aforementioned agent, calling upon different agents to further analyze the tasks.

\item Sub-agent execution: Invoking different sub-agents to perform various visualization tasks.

\item Result output: Presenting the results of sub-agents or the agent itself combined with user input prompts in textual or visual formats.

\item Task evaluation and feedback: Assessing the output results based on user feedback from step three. If the task remains unresolved, returning to steps one through three.
\end{enumerate}

Through this process, the visualization agent can integrate large language models and IoT data temporal-spatial analysis, reducing the threshold of data analysis and enhancing the efficiency of data analysis.

Fig.~\ref{fig:taskExample} illustrates the basic capabilities of the spatiotemporal fusion agent in visualizing data during the process of handling IoT data analysis tasks. In Fig.~\ref{fig:taskExample} (a), the user requests to obtain the location information of all sensor nodes and to retrieve both real and predicted data for a specific node on a particular day. When displaying the positions of sensor nodes, the spatiotemporal fusion agent selects to call upon a sub-agent for drawing scatter plots to execute the task and outputs the corresponding scatter plot. When presenting the predicted values for a single node, the agent distributes the task of drawing line plots to a sub-agent based on the prediction values obtained by the temporal agent analysis and outputs the final results. Fig.~\ref{fig:taskExample} (b) demonstrates the feedback function of the spatiotemporal fusion agent. Initially, the agent displays the similarity relationships computed by the spatial agent as a force-directed graph according to the user's initial request. Subsequently, upon the user's request for further cluster analysis, the spatiotemporal fusion agent visualizes the results of cluster analysis conducted by the spatial agent.

The present spatiotemporal fusion agent consists of sub-agents tasked with visualizing the temporal attributes of IoT data, illustrating the spatial attributes of IoT data, and presenting spatio-temporal fused data. The temporal attribute sub-agent can accomplish tasks such as generating scatter plots of sensor location information and line plots for data predictions. The spatial attribute sub-agent is tasked with producing node clustering diagrams and sensor similarity network graphs. Meanwhile, the spatio-temporal fusion sub-agent is responsible for exhibiting data from different clusters in time post-clustering, while also visualizing data simultaneously in both temporal and spatial dimensions.

\section{Experiments}
To evaluate the framework proposed in this paper, we conducted experimental evaluations on different agent models. The experimental setup was conducted in software environments based on Anaconda for Python 3.10, PyTorch 2.0, and CUDA 11.3.

\subsection{Evaluation of Temporal Agent}
The experiments of the Temporal Agent mainly include the assessment of prediction models and hyperparameter optimization algorithms.

\subsubsection{Performance of Prediction Model} The datasets used in this section are the publicly available CityPulse dataset and the SML2010 dataset~\cite{misc_sml2010_274}. The CityPulse dataset~\cite{ali2015citybench} comprises IoT spatiotemporal data collected in the city of Odense, Denmark, from 2013 to 2016.  For analysis in this section, we selected air pollution data and traffic flow data collected by sensors during August and September 2014. The pollution data includes 17,568 atmospheric pollution readings collected at 217 observation points during these two months at 5-minute intervals, along with vehicle counts and average vehicle speeds recorded by the sensors. The collected pollution data types include ozone, particulate matter, carbon monoxide, sulfur dioxide, and nitrogen dioxide.

The SML2010 dataset is a machine learning dataset provided by UCI. It consists of indoor environmental information collected from environmental sensors installed in residential homes. The data was collected at 15-minute intervals over a span of approximately 40 days. The dataset comprises 24 features, including 20 features related to environmental information such as living room carbon dioxide concentration, indoor relative humidity, and indoor light intensity.

\textbf{Evaluation Metrics}.
In this section of the experiments, three commonly used performance metrics for regression tasks were employed to evaluate the performance of the prediction models. These metrics include 
$RMSE=\sqrt{\frac{1}{N}\sum_{i}(y_i-\hat{y}_i)^2}$, 
$MAE = \frac{1}{N}\sum_{i}|y_i-\hat{y}_i|$, and 
$R^2 = 1 - \frac{\sum_{i}(y_i-\hat{y}_i)^2}{\sum_{i}(y_i-\overline{y})^2}$, where $N$ represents the number of samples, $y_i$ represents the true value of the i-th sample,$\hat{y}_i$ denotes the predicted value by the model, and $\overline{y}$ denotes the mean of the true values.

\textbf{Baselines}. The comparative models used include Multilayer Perceptron (MLP), Gated Recurrent Unit (GRU)~\cite{lai2018modeling}, LSTM~\cite{zhao2017lstm}, Bidirectional Long Short-Term Memory (BiLSTM)~\cite{ma2021short}, Stacked LSTM~\cite{luo2017revisit}, LSTM with attention mechanism~\cite{sun2023time}, GridLSTM~\cite{xu2018single}, and Transformer. MLP and Transformer are non-recurrent neural networks, while the rest are commonly used deep learning models in time-series data analysis tasks. 

\textbf{Configuration.}
The maximum number of iterations for all models, including the proposed model, is set to 200, and the optimizer used is the Adam optimizer provided by PyTorch. The initial learning rate is set to $e^{-3}$. To prevent overfitting, regularization techniques are applied in the models with a weight decay of $e^{-5}$. The loss function used for training is the mean squared error loss function (MSELoss) provided by the PyTorch framework. Each training sample for the models consists of the feature values of the past 12 time steps, and the batch size for model training is set to 200.

\begin{table}[h]
  \centering
  \small
  \caption{Prediction Experiment Results}
  \scalebox{0.8}{
    \begin{tabular}{ccccccc}
    \toprule
    \multirow{3}{*}{Model}&
    \multicolumn{3}{c}{\bf{CityPulse Dataset} }&\multicolumn{3}{c}{\bf{SML2010 Dataset}}\cr
    \cmidrule(lr){2-4} \cmidrule(lr){5-7}
    &RMSE&MAE&$R^2$&RMSE&MAE&$R^2$\cr
    \midrule
    MLP                 &0.2876	&0.2206	&0.9218		&0.0531	&0.2315	&0.9526\cr
    GRU	                &0.0682	&0.0599	&0.9951		&0.0422	&0.1533	&0.9795\cr
    LSTM                &0.0681	&0.0595	&0.9953		&0.0459	&0.1664	&0.9744\cr
    BiLSTM              &0.0675	&0.0594	&0.9952		&0.0386	&0.1518	&0.9808\cr
    Stacked LSTM        &0.0678	&0.0597	&0.9954		&0.0417	&0.1586	&0.9787\cr
    GridLSTM            &0.0677	&0.0594	&\bf{0.9955}&0.0337	&0.1167	&0.9983\cr
    Attentional LSTM    &0.0679	&0.0596	&0.9952		&0.0341	&0.1222	&0.9885\cr
    Transformer         &0.0673	&0.0972	&0.9948		&0.0335	&0.1031	&0.9992\cr
    CityGPT            &\bf{0.0666} &\bf{0.0593} &\bf{0.9955}		&\bf{0.0320}	&\bf{0.0913}	&\bf{0.9994}\cr
    \bottomrule
    \end{tabular}}
    \label{table:prediction_result}
\vspace{-3mm}
\end{table}

The performance metrics of each model on the prediction tasks for the two datasets are presented in Table~\ref{table:prediction_result}. For MLP, a non-temporal model, its performance on the prediction tasks for both datasets is inferior to that of other temporal models. This is attributed to the fact that non-temporal models rely solely on the information available at the current time step for making predictions. In contrast, temporal models can leverage both the information from the current time step and the past time steps for prediction. Results indicate that various variants of LSTM, which have been enhanced, demonstrate superior performance across all performance metrics. Particularly, the Attentional LSTM outperforms other models across all metrics. This phenomenon suggests that the introduction of attention mechanisms in deep learning models enhances feature extraction capabilities, thus improving prediction performance. Comparing LSTM  with GridLSTM, it is observed that GridLSTM, which extracts features from two dimensions, exhibits more prominent capabilities in spatiotemporal data analysis. Furthermore, comparing GridLSTM with CityGPT reveals that incorporating a cross-attention mechanism in the decoder stage to fuse information from both dimensions further enhances the predictive capabilities of the model.

\subsubsection{Hyperparameter Optimization}
The experiment conducted in this section aims to compare the hyperparameter optimization efficacy of PGA employed by us with that of other hyperparameter optimization algorithms concerning the task of data feature prediction using the SML2010 dataset. By documenting the prediction performance and configuration time of each model, a comparison of the performance among different models in hyperparameter configuration tasks is facilitated. Regarding the prediction performance metrics of the models, the metrics utilized in this section align with those employed in the preceding experiments, encompassing RMSE, MAE, and $R^2$. As for the configuration time, it denotes the duration from the initiation of each hyperparameter configuration method to the conclusion of the iteration leading to the derivation of the hyperparameter configuration plan.

\textbf{Baselines}. We compared it with the genetic algorithm-based (GA) automated hyperparameter optimization method{~\cite{young2015optimizing}, manual search, grid search{~\cite{Lerman1980FittingSR}}, Bayesian optimization (BO){~\cite{yu2020hyper}} and Heteroscedastic Evolutionary Bayesian Optimisation (HEBO){~\cite{cowen2020hebo}}.

\textbf{Configuration.}
In manual search, the GridLSTM network has 64 hidden units. For grid search, candidate numbers of neural units per layer are 16, 32, 64, 128, 256, and 512. In other experiments, the number of hidden layer units is determined by each method. BO and HEBO have a solution range from 1 to 512, with Gaussian processes as surrogate models. BO uses 1 point per iteration, while HEBO uses 8 points. GA initializes with a population size of 12. PGA launches 2, 4, and 6 processes successively, dividing sub-populations accordingly. PGA concurrently runs genetic algorithms in multiple processes for automatic hyperparameter optimization. When exchanging between sub-groups, the maximum exchange number is set to 20\% of the population size, with a minimum of 1 exchange. Other hyperparameter configurations for prediction models remain consistent with previous comparative experiments.

\begin{table}[htb]
	\small
	\caption{The Evaluation Metrics with Different Hyperparameter Optimization Methods in SML2010 Dataset}
	\centering
	\label{tab_metrics}
	\setlength{\tabcolsep}{1mm}{
	\begin{tabular}{cccc}
		\toprule
		Methods & \textbf{RMSE} & \textbf{MAPE} & \textbf{R$^2$}  \\ 
            \midrule
		PGA (ours) & {\bf 0.0320} &{\bf 0.1009}&{\bf 0.9997} \\
		GA & 0.0384  & 0.1434 & 0.9996 \\
		  BO & 0.0389 & 0.1314 & 0.9995 \\
		HEBO & 0.0355 & 0.1097 & 0.9996 \\
		Grid Search & 0.0345 & 0.1264 & 0.9997 \\ 
		Manual Optimization & 0.0402 & 0.1525 & 0.9996 \\
	\bottomrule	
	\end{tabular}}
	\label{table:evaluation_SML2010} 
\vspace{-3mm}
\end{table}

Table~\ref{table:evaluation_SML2010} presents the final prediction results of the prediction models on the relevant features of the SML2010 dataset under different hyperparameter configuration algorithms. Overall, hyperparameter configuration methods based on PGA outperform other algorithms. From RMSE and MAE, PGA and HEBO perform well, indicating that these two hyperparameter configuration methods are suitable for optimizing deep learning models. Regarding $R^2$ , the performance of various algorithms is similar, with only PGA and BO showing a slight advantage. This is because $R^2$ reflects the extent to which the dependent variable is determined by the independent variable, focusing on the correlation between the two, and the input features of the comparative experimental models are consistent. Therefore, the performance of each experiment on this indicator is similar.

\vspace{-2mm}
\begin{figure}[htb]
    \centering
    \includegraphics[width=0.7\linewidth]{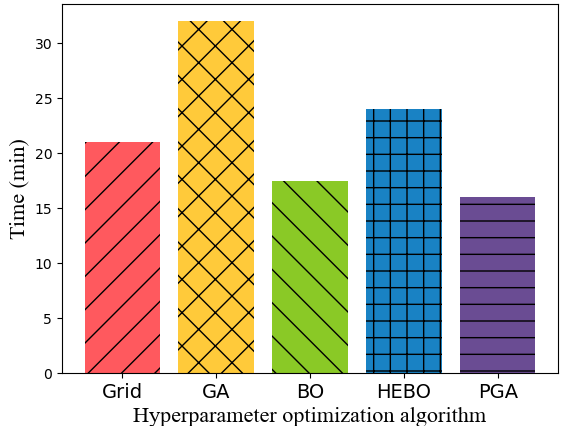} 
    \vspace{-3mm}
    \caption{Running time of each model} 
    \vspace{-3mm}
    \label{fig:HPO_time} 
\end{figure}

Fig.~\ref{fig:HPO_time} illustrates the time taken for each configuration model. Manual parameter tuning time is not included due to difficulty in accurate tracking. PGA utilized 2 processes in the figure. Both PGA and BO had the shortest runtime, with PGA slightly outperforming BO. Comparing GA with PGA, GA's runtime was nearly double that of PGA due to PGA's multi-process environment, leveraging parallelism for faster runtime. However, PGA involves additional inter-process data exchange, resulting in a slightly longer runtime, approximately half of GA's. HEBO's runtime was significantly longer than BO's due to computing the effectiveness of eight hyperparameter configurations per iteration, increasing computational load. Grid search had a relatively short runtime as only one hyperparameter was optimized. If multiple hyperparameters were optimized, grid search's computational load would increase, potentially exceeding GA's computation time.

\subsection{Evaluation of Spatial Agent}
To validate Cross-Graph CosimHeat's effectiveness within the Spatial Agent, we applied it to the CityPulse dataset for pollution data analysis, emphasizing inter-regional correlations.

\subsubsection{Sensor Spatial Distribution}
The CityPulse dataset comprises data collected from 217 sensors, with the geographic information of each sensor summarized in a table in terms of latitude and longitude. Based on the geographical locations of these sensors, they are plotted on a map to provide an initial visualization of the distribution of data collection points in the area. Fig.~\ref{fig:SensorMap} (a) illustrates the overall distribution of all sensors in the Aarhus region. It is observed from this figure that most sensors are concentrated in the urban areas of the region, with fewer sensors located in suburban or rural areas. Fig~\ref{fig:SensorMap} (b) presents the positions of sensors on a smaller scale. As depicted in this figure, the sensors are positioned along the roads.

\begin{figure}[h]
\begin{center}
\begin{tabular}{cc}
\includegraphics[width=0.4\linewidth]{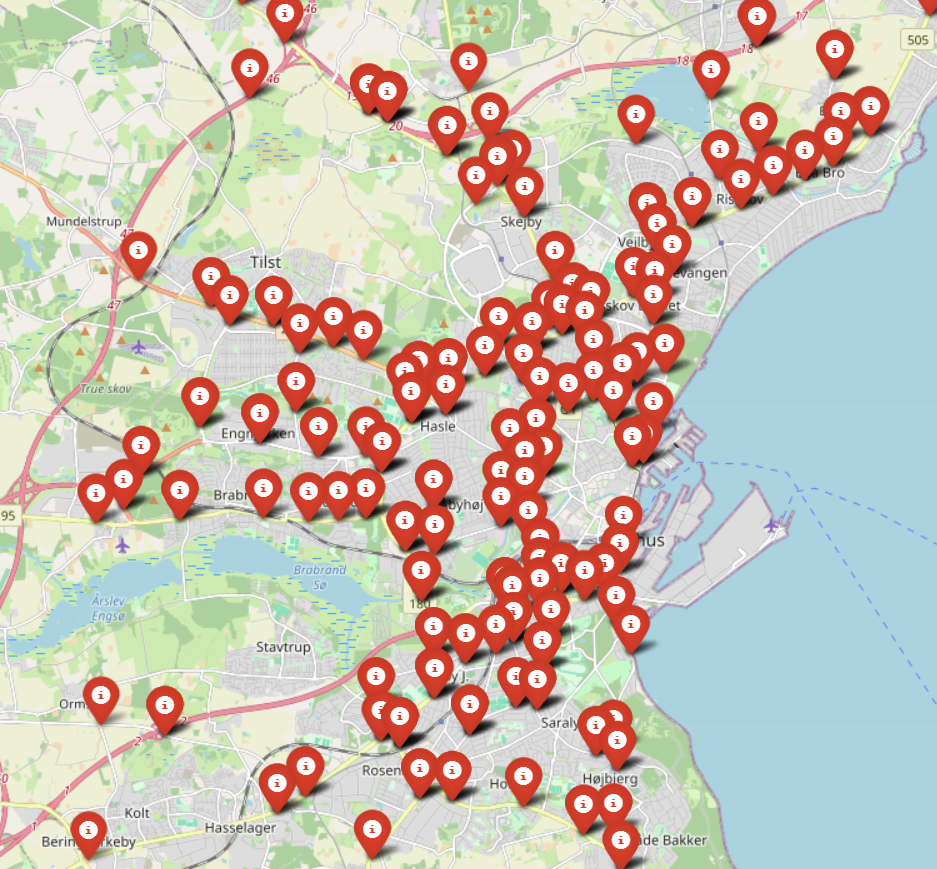} & \includegraphics[width=0.45\linewidth]{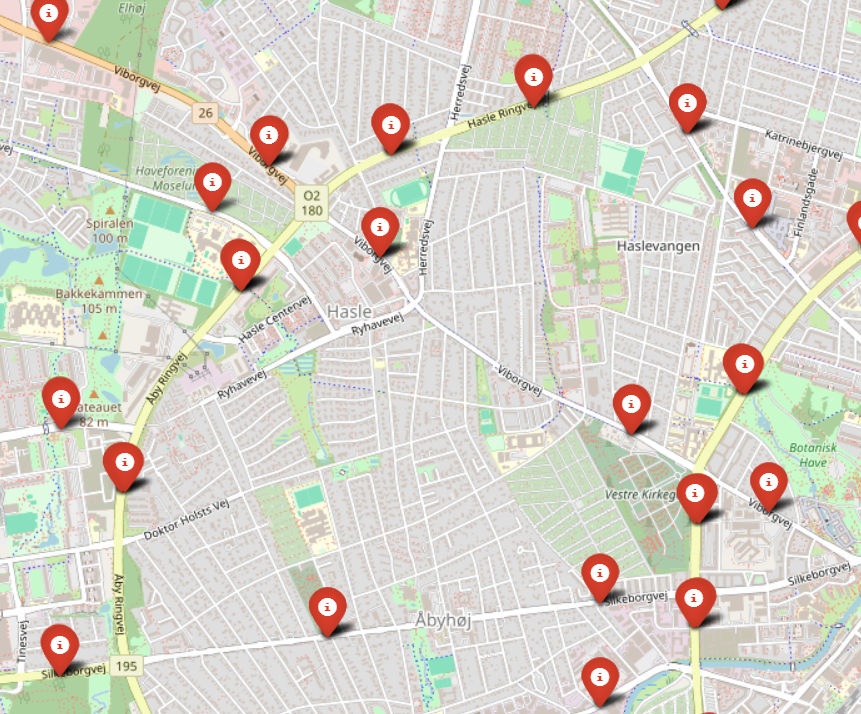} \\
{\scriptsize(a) Overall Distribution} & {\scriptsize (b) Local Distribution}
\end{tabular}
\caption{Sensor Distribution Map}
\vspace{-3mm}
\label{fig:SensorMap}
\end{center}
\end{figure}

\subsubsection{Graph Construction and Similarity Calculation}By treating the region covered by the CityPulse dataset as a graph and the sensors in the dataset as nodes in the graph, an abstract graph structure of the area is constructed. Each node's representation in the feature space is formed by a multidimensional vector composed of different data features. Then, the initial similarity relationship between nodes is established by computing the representation vectors of pairwise nodes, resulting in the construction of the initial similarity matrix for the graph structure of the area. Subsequently, leveraging CoSimHeat on this matrix computes the final similarity matrix results for each node. The initial similarity between nodes is defined as Eq.~\eqref{equ:Cosim15}.

\begin{equation}
\label{equ:Cosim15}
    Similarity(i,j) = \alpha \cdot dis(i,j) + (1 - \alpha) \cdot Cos(\boldsymbol{X}_i, \boldsymbol{X}_j)
\end{equation}

\noindent
where $i$ and $j$ represent node $i$ and node $j$. $dis(i,j)$ characterizes the spatial similarity between the two nodes, which denotes the geographical distance between nodes $i$ and $j$. $Cos(\boldsymbol{X}_i, \boldsymbol{X}_j)$ represents the cosine similarity value between the representation vectors of these two nodes, indicating the similarity in terms of data feature dimensions. 
$\alpha$ is a parameter used to control the influence of these two features on the similarity between nodes, with its value ranging between 0 and 1.

In this experiment, we set $\alpha = 0.5$, indicating an equal weighting between the two types of features. Additionally, due to the cosine similarity values falling within the range of $[0,1]$, the value of $Cos(\boldsymbol{X}_i, \boldsymbol{X}_j)$ is normalized during the actual computation process, as shown in Eq.~\eqref{equ:Cosim16}.

\begin{equation}
\label{equ:Cosim16}
    dis(i,j)=1-\frac{g_{i,j}-g_{min}}{g_{max}-g_{min}}
\end{equation}

\noindent
where $g_{i,j}$ represents the actual geographical distance between nodes $i$ and $j$. $g_{min}$ and $g_{max}$ are the minimum and maximum distance values among all pairs of nodes.


\begin{figure}[htb]
	\centering
	\includegraphics[width=0.7\linewidth]{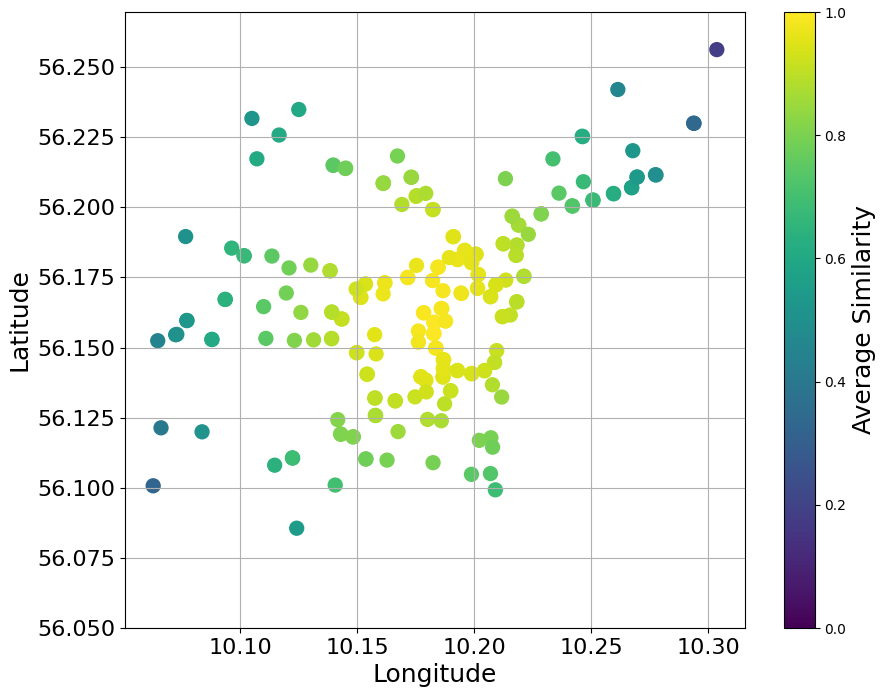} 
        \vspace{-3mm}
	\caption{Average similarity heat map} 
        \vspace{-3mm}
	\label{fig:averageSimilarity} 
\end{figure}

Fig.~\ref{fig:averageSimilarity} presents the visualization results of the averaged similarity matrix computed with CoSimHeat. Brighter node colors indicate higher average similarity with all other nodes, reflecting stronger connections in vehicle movement. Conversely, darker colors denote lower average similarity. Spatial distribution patterns of node average similarity are evident. Nodes with higher average similarity are concentrated in a grid-like area with longitude between 10.15 and 10.20 and latitude between 56.14 and 56.20. Surrounding this central area, nodes in adjacent regions show lower average similarity compared to others. Bright areas suggest higher similarity in vehicle flow, indicating bustling urban areas with larger traffic volumes. Conversely, similarity gradually decreases outward, signifying relatively marginalized areas with lower traffic flow. By comparing with Fig.~\ref{fig:SensorMap}, brighter regions correspond to densely populated urban areas, while darker regions indicate suburban areas with higher greenery levels.

\subsubsection{Spectral Cluster Analysis}
Based on the node similarity matrix, spectral clustering is employed to cluster the nodes, with the number of clusters varying from 2 to 8. Visualizations of clustering results for 4, 5, 6, and 7 categories are shown in Fig.~\ref{fig:clusters}. Comparing Fig.~\ref{fig:clusters} (a) to Fig.~\ref{fig:clusters} (b), a new category emerges from the partitioning of categories 0 and 3. Similarly, comparing Fig.~\ref{fig:clusters} (b) to Fig.~\ref{fig:clusters} (c), the newly added category mainly originates from the partitioning of category 0. Further comparison between Fig.~\ref{fig:clusters} (c) and Fig.~\ref{fig:clusters} (d) reveals that the new category arises from the partitioning of category 2. Despite changes in some categories with increased clustering categories, certain node clusters remain stable, particularly in the geographic area with longitude ranging from 10.17 to 10.22 and latitude ranging from 56.13 to 56.17. Additionally, the classification of nodes in the upper right corner of the geographic area remains consistent, indicating stable spatial similarity and vehicular traffic patterns.


\begin{figure}[tb]
	\begin{center}
		\begin{tabular}{cc}
			\includegraphics[width=0.47\columnwidth]{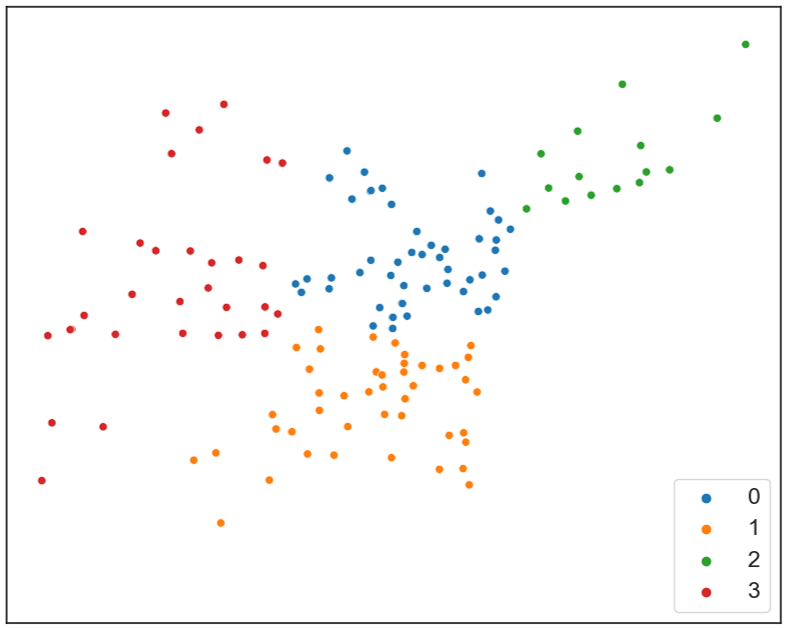}&
			\includegraphics[width=0.47\columnwidth]{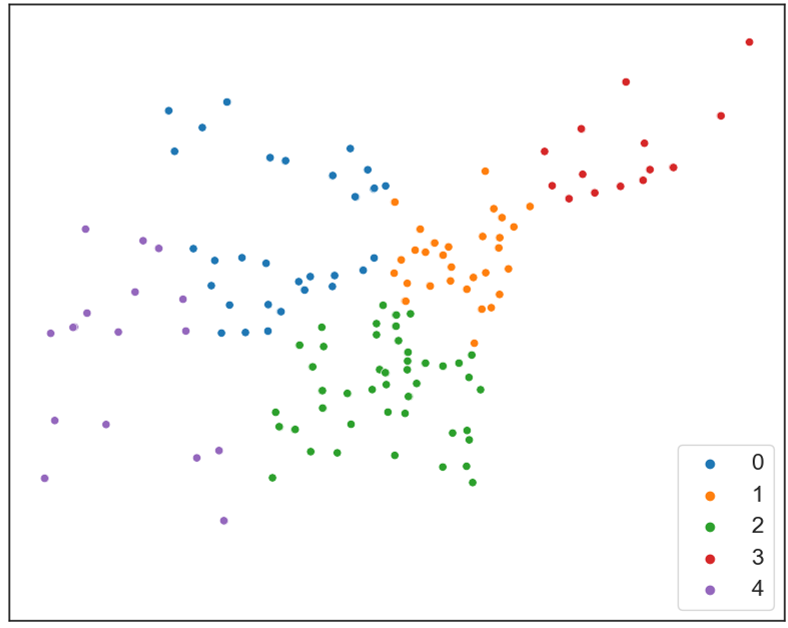}\\
            {\scriptsize (a) The number of clusters is 4}& {\scriptsize (b) The number of clusters is 5}\\ 
            \includegraphics[width=0.47\columnwidth]{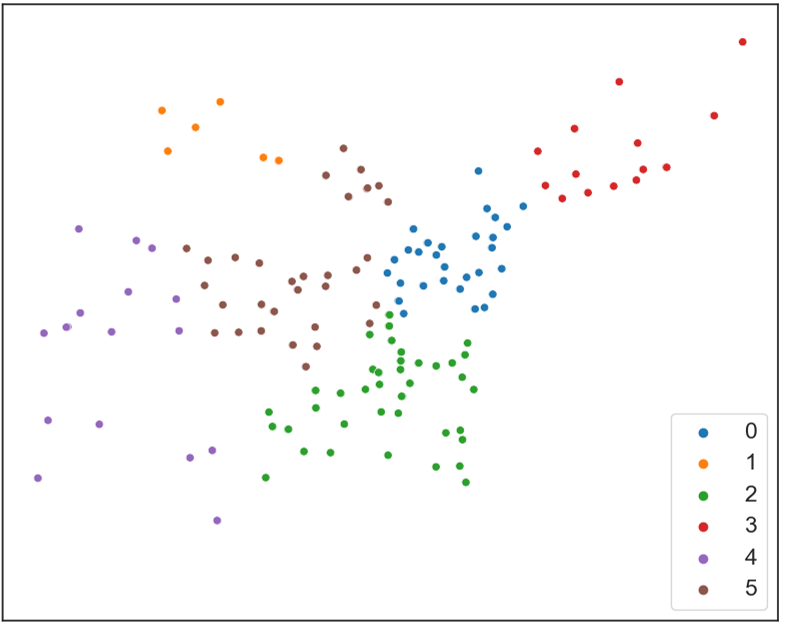} & \includegraphics[width=0.47\columnwidth]{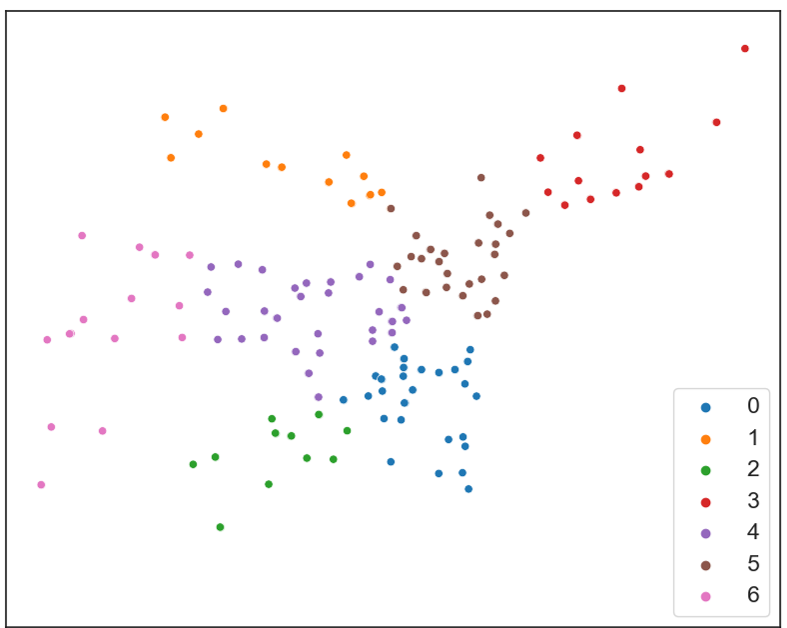}\\
			{\scriptsize (c) The number of clusters is 6} & {\scriptsize (d) The number of clusters is 7}
		\end{tabular}
	\end{center}
        \vspace{-3mm}
	\caption{Results for different number of clusters}
        \vspace{-3mm}
	\label{fig:clusters}
\end{figure}

Table~\ref{table:cluster_result} presents the values of three commonly used clustering performance metrics, the Silhouette Coefficient (SC) index, Calinski-Harabaz (CH) index, and Davies-Bouldin (DB), for spectral clustering results under different numbers of clusters. When the number of clusters is set to 3, spectral clustering exhibits the most favorable SC index. However, when the number of clusters is 5, the algorithm demonstrates optimal performance in terms of the CH index and DB index. Furthermore, when the number of clusters is 5, the silhouette coefficient of spectral clustering is only slightly lower than that when the number of clusters is 3. Considering these factors comprehensively, this study selects the spectral clustering results with 5 clusters as the basis for subsequent data analysis.

\begin{table}[h]
	\small
	\caption{Clustering Evaluation Results}
	\centering
	\label{tab_metrics}
	\setlength{\tabcolsep}{1mm}{
	\begin{tabular}{cccc}
		\toprule
		cluster number & SC index & CH index & DB index  \\ 
            \midrule
        2	&0.277	&21.744	&1.268\\
        3	&\textbf{0.328}	&23.017	&1.870\\
        4	&0.315	&25.893	&2.044\\
        5	&0.316	&\textbf{40.797} &\textbf{1.317}\\
        6	&0.172	&36.024	&1.568\\
        7	&0.188	&33.515	&1.468\\
        8	&0.190	&29.213	&1.521\\

	\bottomrule	
	\end{tabular}}
	\label{table:cluster_result} 
\vspace{-3mm}
\end{table}

\subsubsection{Comparative Analysis between Clusters}

To assess the differences between clusters quantitatively, this study utilized the Cross-Graph CoSimHeat outlined in Section~\ref{sect:similarityModel} to compute the similarity between each pair of clusters. A refined graph structure was constructed using road traffic network data from OpenStreetMap in the Aarhus area. Nodes within the same street in the same cluster were considered connected when constructing the adjacency matrix. Each node was associated with a value vector, similar to the representation vectors mentioned earlier. Edge weights between nodes were defined as the cosine similarity of their value vectors. If nodes from different clusters were located on the same street, they were considered connected, with edge weights calculated similarly based on cosine similarity. This initialization provides all necessary data for the Cross-Graph CoSimHeat.

\begin{figure}[h]
\begin{center}
\begin{tabular}{cc}
\includegraphics[width=0.47\linewidth]{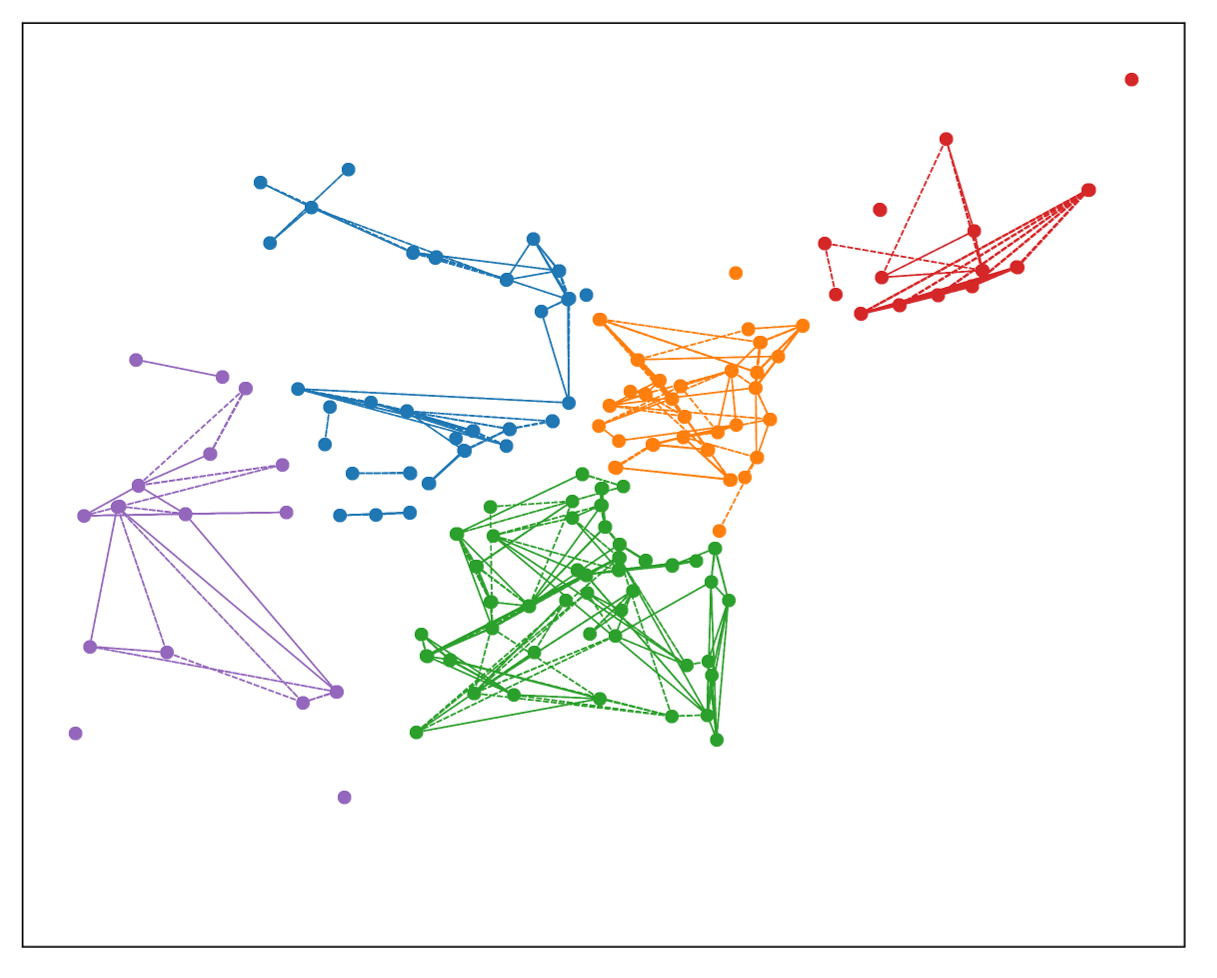} & \includegraphics[width=0.48\linewidth]{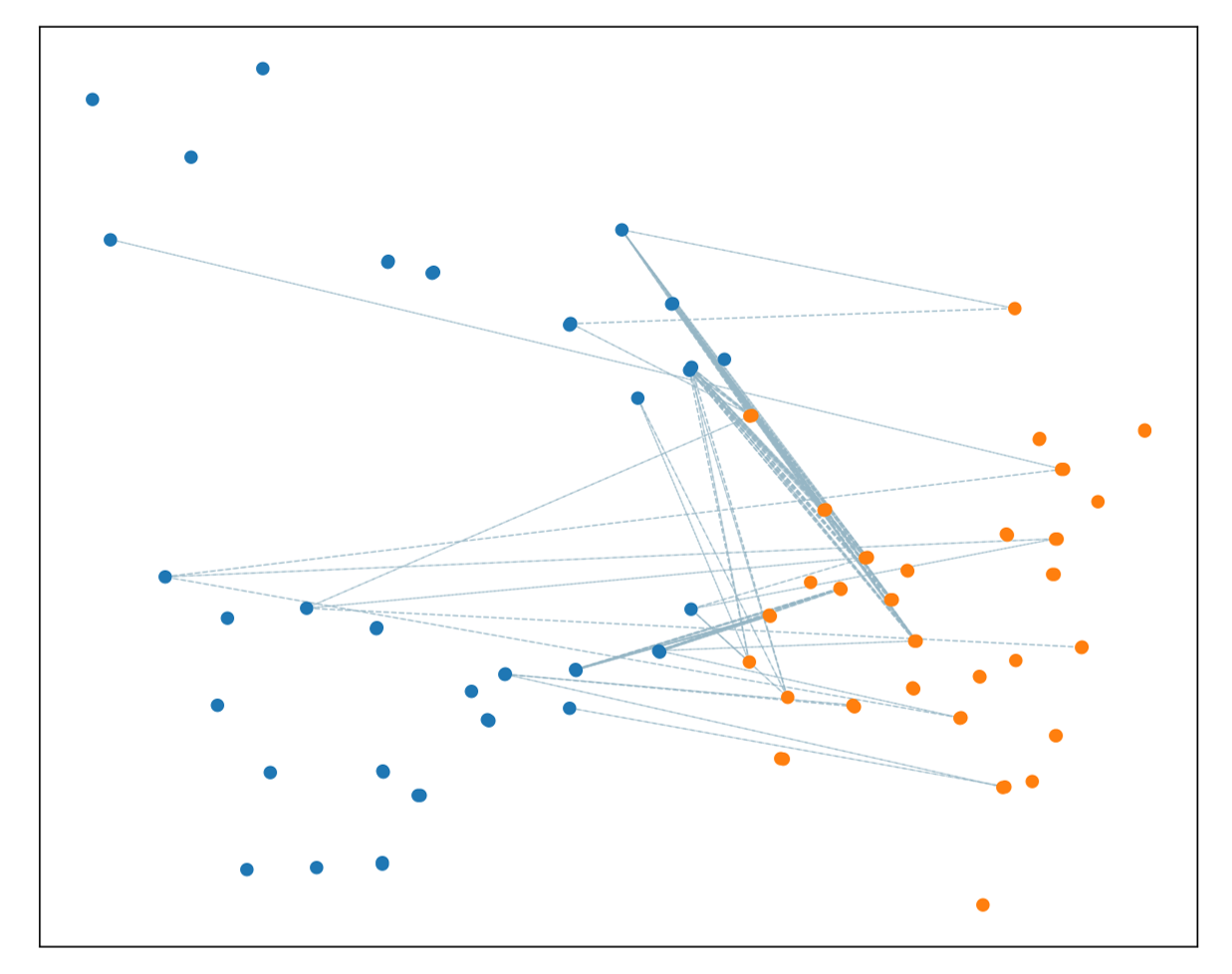} \\
{\scriptsize(a) Edges inside clusters} & {\scriptsize (b) Edges between cluster 0 and cluster 1}
\end{tabular}
    \vspace{-3mm}
    \caption{Edges between clusters}
    \vspace{-3mm}
\label{fig:Edge}
\end{center}
\end{figure}

Fig.~\ref{fig:Edge} (a) delineates the intra-cluster edge relationships, while Fig.~\ref{fig:Edge} (b) elucidates the preliminary inter-cluster node associations specifically between clusters 0 and 1. Within both representations, nodes allocated to distinct clusters are visually distinguished by diverse-colored scatter points. The interconnecting lines between scatter points symbolize the presence of edges linking the respective nodes, thereby signifying their adjacency along shared thoroughfares. Notably, in Fig.~\ref{fig:Edge} (a), certain nodes may exhibit isolation owing to their remote positioning, wherein they lack shared thoroughfares with other nodes. Correspondingly, in Fig.~\ref{fig:Edge} (b), certain nodes within cluster 0 may lack direct edge connections with nodes in cluster 1. Nonetheless, leveraging the Cross-Graph CoSimHeat facilitates the inference of similarity relationships among all nodes spanning across the two delineated clusters.


\begin{figure}[htb]
	\centering
	\includegraphics[width=0.7\linewidth]{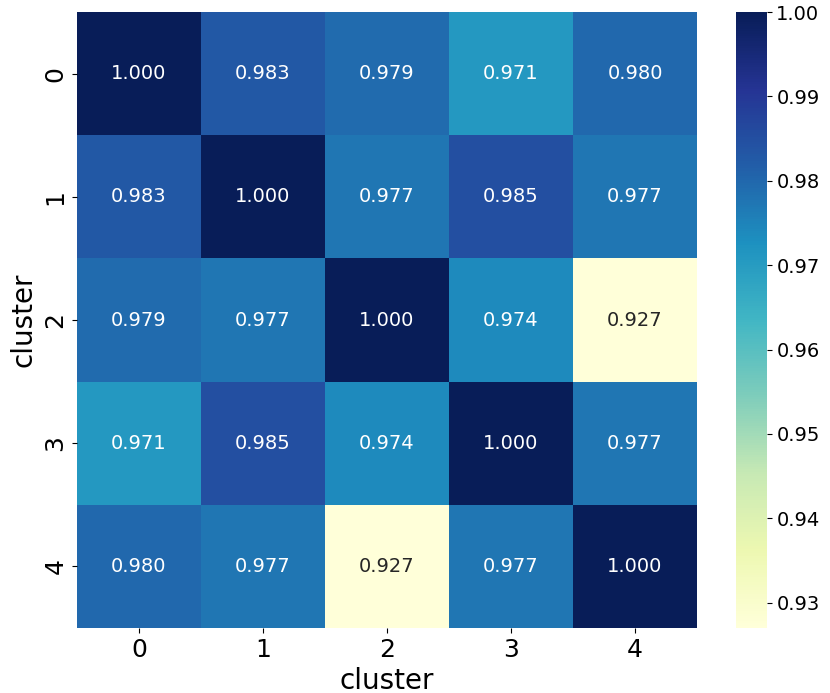} 
        \vspace{-3mm}
        \caption{Similarity between clusters} 
	\label{fig:similarityHeat} 
        \vspace{-3mm}
\end{figure}

\begin{figure*}[tb]
    \centering
    \includegraphics[width=\linewidth]{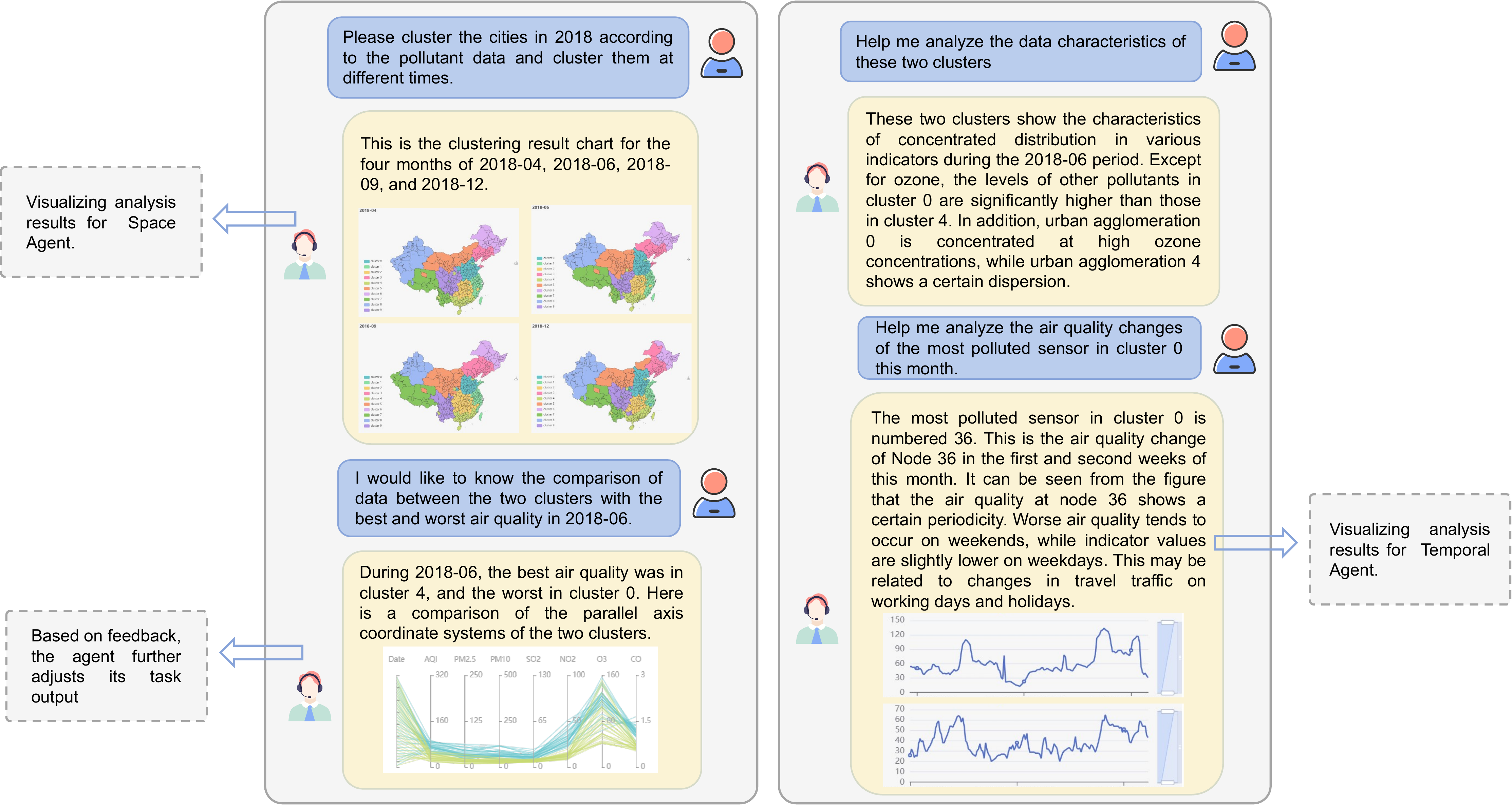}
    \vspace{-3mm}
    \caption{Analysis of air pollution data by CityGPT} 
    \vspace{-3mm}
    \label{fig:Case_study} 
\end{figure*}


After obtaining adjacency matrices for each cluster and initial cluster relationships, we computed the similarity matrix between clusters using Eq.\eqref{equ:Cosim14}. To gauge the overall similarity between clusters, we summed all elements in the similarity matrix, providing a measure of their overall similarity. With nodes classified into five categories through spectral clustering, we conducted inter-cluster similarity calculations among these five clusters, resulting in 10 comparisons. The heatmap in Fig.~\ref{fig:similarityHeat} visually represents the overall similarity relationships among clusters, with darker colors indicating higher similarity and lighter colors indicating lower similarity. Each cluster exhibits perfect self-similarity, while variations exist in inter-cluster similarities. For example, clusters 0 and 1 demonstrate a high similarity score of 0.983, indicating similar node features. Conversely, cluster 4 shows a lower similarity score of 0.927 than cluster 2, suggesting significant differences in pollutant values. Urban clusters (0, 1, 2) display higher similarity, reflecting consistent pollutant concentration values. This analysis identifies regions with similar environmental characteristics and underscores differences in pollutant distribution among areas.

\subsection{Case Study}
This section demonstrates how the system in this paper conducts data analysis and visualization with a specific case study. Fig.~\ref{fig:Case_study} shows the analysis of air pollution data across several Chinese cities using the proposed agent, involving four interaction rounds.

In the first round, the user requests clustering analysis of the data from a spatial perspective. Upon receiving the command, the agent computes the similarity between these cities through spatial representation and subsequently applies spectral clustering to categorize them. The clustering results are then transmitted to the spatio-temporal fusion agent. Upon receiving the visualization task, the spatio-temporal fusion agent inputs the clustering results into the clustering visualization sub-agent and presents the city clustering graphs for the months of April 2018, June 2018, September 2018, and December 2018 to the user. In the second and third rounds of interaction, the user provides feedback and requests further comparative analysis of the data for June 2018. In response, the system conducts a comparative analysis of the regions represented by different clusters over time using parallel coordinate plots and provides corresponding analytical descriptions: "Except for ozone, the levels of other pollutants in cluster 0 are significantly higher than those in cluster 4. In addition, urban agglomeration 0 is concentrated at high ozone concentrations, while urban agglomeration 4 shows a certain dispersion." In the fourth round, the user requests the display of data for specific nodes within a cluster. At this point, the system visualizes the data through the time visualization sub-agent within the spatio-temporal fusion agent and provides the corresponding data analysis results.

These four rounds of dialogue demonstrate that the system can allocate tasks according to user input execute tasks, and present results using different modules of agents. Additionally, the third round of dialogue showcases that the system can modify tasks based on user feedback to produce the desired results.

\section{Related Work}
\subsection{Spatiotemporal Data Processing}
\textbf{LSTM variants.} RNNs~\cite{SaeedUBICOMP2019} can handle long-delayed tasks without predefined time steps, but often face gradient issues in longer time series. Long Short-Term Memory (LSTM) networks~\cite{hochreiter1997long}, an RNN variant, solve this with a gating mechanism for memory cells. However, single LSTMs only retain memory in the time direction and cannot process complex features. Stacked LSTMs~\cite{WuTII2019} and Bi-Directional LSTMs~\cite{Schuster1997Bidirectional} improve this by stacking layers and processing data forward and backward, respectively, enabling longer-range patterns. Yet, they lack multidimensional feature learning due to limited vertical computation. Google DeepMind's Grid LSTM extends LSTM cells along multiple dimensions, including network depth, enhancing performance for tasks like single-channel speech separation~\cite{andrea2018vector}. For IoT-specific features, adaptive changes to the Grid LSTM mechanism are needed to effectively learn time series data from various sensors.

\noindent
\textbf{Graph similarity computation.} Graph similarity computation is vital in graph theory, aiming to gauge the similarity between two graphs. The core concept involves computing similarity metrics by comparing their structures or other attributes. The main methods for graph similarity computation are graph kernel-based and graph matching-based. Graph kernel-based methods transform non-linearly separable problems in low-dimensional space into linearly separable problems in high-dimensional space. These methods encompass path-based graph kernels~\cite{huang2021lcs}, subgraph-based graph kernels~\cite{wang2022towards}, and subtree-based graph kernels~\cite{schulz2022generalized}. On the other hand, graph matching-based similarity computation maps one graph onto another and calculates similarity by comparing the degree of correspondence between the mapped graph and the original graph. This approach often involves techniques such as graph isomorphism matching and maximum common subgraph matching, making it suitable for handling correspondences between two graphs~\cite{fuchs2021matching, zeng2018learning, li2022sigma}.

\subsection{Hyperparameter Optimization}
In order to realize the automatic setting of multiple steps in machine learning, AutoML has been proposed and has become a research hotspot in academia and industry. Thakur et al. ~\cite{Thakur2015AutoCompeteAF} proposed the AutoCompete framework for machine learning competitions. The framework tries to build a predictive model through learning without human intervention. Liang et al.~\cite{Liang2019GECCO} proposed LEAF, an AutoML framework. The latest evolutionary algorithms and distributed computing frameworks are used simultaneously to optimize hyperparameters, network structure and network size. HPO, essential in AutoML, includes methods like grid search, random search, BO, and gradient-based optimization ~\cite{garg2021mofit,ye2023application,bassi2021comparative}. Recently, some researchers have also started to use genetic algorithms to implement hyperparametric optimization. ~\cite{aszemi2019hyperparameter} presents a genetic algorithm method for optimizing the CNN hyperparameter.
With the need for time-sensitive and simultaneous training, parallel processing is expected but rarely tackled in AutoML.

\subsection{AI Agent with Human-AI Interaction}
Artificial intelligence (AI) agents, as a rising technological framework, have facilitated the transition of system building from the initial data flow research phase to the application phase of big data theory and analysis~\cite{wangimwut2020, diimwut2022}. This revolutionary advancement has enabled the IoT field to gain unheard of growth potential to enhance the manageability of complex Big Data systems for ordinary people.

Several studies have revealed the potential of using AI agents to optimize big data systems. For example, Tang et al.~\cite{Yiqingdtpi2023} used ChatGPT to model a large number of road traffic policies and found that using ChatGPT for traffic policy optimization and increasing the average speed of vehicles. Villarreal et al.~\cite{Michaelitsc2023} found that using a large amount of cross-domain expertise to train a reinforcement learning-related agent can enable novice people to solve complex mixed flow control problems, which typically cover an extremely large amount of data and span a wide range of domains.

However, in the IoT field, the prominent problem is the huge volume of data and the complex structure of spatiotemporal data, which makes it difficult for unusual AI agent systems to interpret it in a generalized way. IoT data has a very high potential for application, but the data volume and domain span prevent it from being widely applied. Simple data pattern analysis will lose a lot of potentially useful information. Therefore, as a general concern, enhancing the interpretability and generalized understanding of IoT data has important implications.

\section{Conclusion}

This paper presents CityGPT for IoT data analysis, facilitating natural language interaction between users and the system to aid in task completion. Initially, a requirement analysis agent translates user inputs into actionable tasks. Leveraging the reasoning capabilities of the large language model, tasks are decomposed into subtasks for execution by individual system agents. Intermediary agents, including temporal and spatial agents, analyze IoT data from distinct temporal and spatial perspectives. The spatiotemporal fusion agent amalgamates results from these agents and inputs them into a visualization sub-agent, presenting visualizations in various formats and providing corresponding textual analysis results. Experimental validation of the system's primary proxies is conducted to verify its reliability, along with showcasing the workflow and effectiveness through a case study. Future efforts will involve integrating additional agents for handling more complex tasks.


\bibliographystyle{IEEEtran}
\bibliography{ARXIV_TRAFF}

\end{document}